\documentclass{article}

    \PassOptionsToPackage{numbers, compress}{natbib}

    \usepackage[final]{neurips_2024}

\usepackage[utf8]{inputenc} 
\usepackage[T1]{fontenc}    
\usepackage[pagebackref=true,breaklinks=true,letterpaper=true,colorlinks,bookmarks=false,citecolor=blue]{hyperref}
\usepackage{url}            
\usepackage{booktabs}       
\usepackage{amsfonts}       
\usepackage{nicefrac}       
\usepackage{microtype}      
\usepackage{xcolor}         
\usepackage{bm}
\usepackage{multirow}
\usepackage{graphicx}
\usepackage{amssymb}
\usepackage{amsmath}
\usepackage{bbding}
\usepackage{pifont}
\usepackage{utfsym}
\usepackage{wrapfig}

\title{\raisebox{-0.2\height}{\includegraphics[width=0.06\textwidth,keepaspectratio]{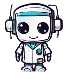}} PediatricsGPT: Large Language Models as Chinese Medical Assistants for Pediatric Applications}

\author{%
  Dingkang Yang$^{1,3 \,\usym{1F396}}$\footnotemark[2]   $\quad$  Jinjie Wei$^{1,3}$\footnotemark[2]
  $\quad$  Dongling Xiao$^{2}$\footnotemark[2] $\quad$ Shunli Wang$^{1}$\footnotemark[4]
  $\quad$ Tong Wu$^{2}$\footnotemark[4] \\ 
  $\quad$ \textbf{Gang Li}$^{2}$\footnotemark[4]  $\quad$ \textbf{Mingcheng Li}$^{1}$\footnotemark[4]
  $\quad$ \textbf{Shuaibing Wang}$^{1}$\footnotemark[4] $\quad$ \textbf{Jiawei Chen}$^{1}$\footnotemark[4] $\quad$ \textbf{Yue Jiang}$^{1}$\footnotemark[4] \\
  $\quad$ \textbf{Qingyao Xu}$^{1}$\footnotemark[4] $\quad$ \textbf{Ke Li}$^{2}$\footnotemark[4] $\quad$ \textbf{Peng Zhai}$^{1,3}$\footnotemark[1]
  $\quad$ \textbf{Lihua Zhang}$^{1,3,4,5}$\footnotemark[1] \\
  \\
 \small $^{1}$Academy for Engineering and Technology, Fudan University, Shanghai, China  \\
 \small $^{2}$Tencent Youtu Lab, Shanghai, China  \\
  \small $^{3}$Cognition and Intelligent Technology Laboratory, Shanghai, China\\
  \small $^{4}$Engineering Research Center of AI and Robotics, Ministry of Education, Shanghai, China\\
  \small $^{5}$AI and Unmanned Systems Engineering Research Center of Jilin Province, Changchun, China\\
 \small \texttt{$\{$dkyang20, pzhai, lihuazhang$\}$@fudan.edu.cn}    \\
  \small \texttt{jjwei23@m.fudan.edu.cn, xdluestc@outlook.com}    \\
    \small \texttt{tristanli@tencent.com}    \\
}

\begin{document}

\maketitle

\renewcommand{\thefootnote}{\fnsymbol{footnote}} 
\footnotetext[2]{Equal first contributions. $^{\S}$Equal second contributions.  $^\ast$Corresponding authors. $^{\usym{1F396}}$Project lead.}  

\begin{abstract}
Developing intelligent pediatric consultation systems offers promising prospects for improving diagnostic efficiency, especially in China, where healthcare resources are scarce.
Despite recent advances in Large Language Models (LLMs) for Chinese medicine, their performance is sub-optimal in pediatric applications due to inadequate instruction data and vulnerable training procedures.
To address the above issues, this paper builds PedCorpus, a high-quality dataset of over 300,000 multi-task instructions from pediatric textbooks, guidelines, and knowledge graph resources to fulfil diverse diagnostic demands.
Upon well-designed PedCorpus, we propose PediatricsGPT, the first Chinese pediatric LLM assistant built on a systematic and robust training pipeline.
In the continuous pre-training phase, we introduce a hybrid instruction pre-training mechanism to mitigate the internal-injected knowledge inconsistency of LLMs for medical domain adaptation.
Immediately, the full-parameter Supervised Fine-Tuning (SFT) is utilized to incorporate the general medical knowledge schema into the models.
After that, we devise a direct following preference optimization to enhance the generation of pediatrician-like humanistic responses.
In the parameter-efficient secondary SFT phase,
a mixture of universal-specific experts strategy is presented to resolve the competency conflict between medical generalist and pediatric expertise mastery.
Extensive results based on the metrics, GPT-4, and doctor evaluations on distinct downstream tasks show that PediatricsGPT consistently outperforms previous Chinese medical LLMs.
The project and data will be released at~\url{https://github.com/ydk122024/PediatricsGPT}.

\end{abstract}

\section{Introduction}

As an essential component of medicine, pediatrics plays an indispensable role in ensuring children’s health growth~\cite{elzouki2011textbook,garrison1923history}.
The unbalanced distribution of healthcare resources~\cite{pu2021fairness} has resulted in a massive shortage of pediatricians, especially in populous countries led by China~\cite{qin2014economic,dong2021differences}.
With the rapid advances in LLMs exemplified by ChatGPT~\cite{ChatGPT}, developing intelligent pediatric consultation systems provides promise for enriching medical services.
Although Chinese LLMs~\cite{cui2023efficient,zhang2022fengshenbang,Baichuan,zeng2022glm,du2021glm} have exhibited progress in general language understanding, they are incompetent in the pediatric medical field due to the lack of domain-specific discipline and specialized expertise injection.

To fulfil the interactive demands of Chinese medicine, preliminary efforts~\cite{bao2023disc,wang2023huatuo,xiong2023doctorglm,chen2023bianque} have enhanced LLMs' healthcare mastery through Supervised Fine-Tuning (SFT) training and medically relevant corpus collection.
Despite improvements, challenges remain due to unavoidable dilemmas, including inadequate instruction data and vulnerable training procedures.
Specifically, (i) existing instruction data typically involve vanilla rephrasing of the general medical corpus~\cite{xiong2023doctorglm} or aggregation of doctor-like dialogues~\cite{yang2024zhongjing}, which loses the specialization and focus in pediatric applications.
More importantly, the current straightforward different round instruction construction paradigms~\cite{zhang2023huatuogpt,chen2023bianque} fail to accommodate multi-task healthcare services in real-world scenarios, limiting the model generalization and inducing response hallucination. 
(ii) Furthermore, prior methods mostly relied on SFT to compensate for medical instruction following capabilities, ignoring the discrepancies between inherent and externally absorbed knowledge within the models.
This single pattern causes secondary LLMs to lapse into excessive role-playing rather than understanding~\cite{ramamurthy2022reinforcement}.
Despite a few attempts in the pre-training and Reinforcement Learning from Human Feedback (RLHF) phases~\cite{bai2022training,ouyang2022training}, their performance is restricted by actor-critic instability~\cite{schulman2017proximal} and online sampling bias~\cite{zheng2023secrets}.

Motivated by these observations, we construct PedCorpus, a high-quality dataset with over 300,000 instructions across single-turn and multi-turn medical conversations.
Besides containing generalist healthcare data, PedCorpus incorporates multi-dimensional corpora from pediatric textbooks, guidelines, and knowledge graphs to ensure medical knowledge's accuracy.
Vanilla instructions can also be readily extended to seed instructions for generating specialized corpora to serve different training phases.
Furthermore, we integrate the well-presented GPT-4-distilled data with authentic doctor-patient dialogue data to standardize the fluency and faithfulness of instruction information.

Among our PedCorpus, we propose PediatricsGPT, the first Chinese pediatric LLM assistant with pediatric expertise and medical generalist.
PediatricsGPT is developed on a systematic training pipeline that includes Continuous Pre-Training (CPT), full-parameter SFT, human preference alignment, and parameter-efficient secondary SFT.
In this case, we introduce a hybrid instruction pre-training mechanism in CPT to bridge the capability weakening due to corpus format discrepancies between the internal and injected medical knowledge of foundation models, facilitating knowledge accumulation and extension.
Meanwhile, a Direct Following Preference Optimization (DFPO) in human preference alignment is devised to enhance response robustness and align human preferences.
Additionally, we present a mixture of universal-specific experts strategy to tackle the competency conflict between medical generalist and pediatric expertise in secondary SFT via Low-Rank Adaptation (LoRA)~\cite{hu2021lora}, which strengthens the model's adaptability to distinct downstream tasks.
We conduct three pragmatic pediatric tasks to evaluate the different capabilities of existing models.
Extensive experiments on pediatric and public benchmarks show that our PediatricsGPT outperforms open-source Chinese medical LLMs and baselines, yielding competitive performance compared to GPT-3.5-turbo.

\vspace{-5pt}
\section{Related Work}
\vspace{-5pt}
\textbf{Chinese Large Language Model Evolution.}
The emergence of Large Language Models (LLMs) dominated by ChatGPT~\cite{ChatGPT} and GPT-4~\cite{achiam2023gpt} has revolutionized the paradigm for novel human-machine interaction. Driven by learning-oriented technologies~\cite{chen2024can,chen2024miss,chen2024efficiency,yang2022disentangled,yang2022learning,yang2024robust,wei2024medaide}, pragmatic instruction~\cite{mishra2021natural,wei2021finetuned} and preference optimization~\cite{bai2022training,ouyang2022training} strategies enable LLMs to address complex generation tasks with aligned human intentions.
Despite improvements, large-scale resources for training general LLMs~\cite{jiang2023mistral,touvron2023llama,touvron2023llama-2} are anchored in the English corpora, limiting their abilities to respond reliably in extensive Chinese application scenarios.
Recently, researchers~\cite{cui2023efficient,zhang2022fengshenbang} have attempted to enhance the comprehension and execution of Chinese instructions in open-source LLMs by augmenting Chinese vocabulary and data (\textit{e.g.,} Chinese LLaMA and Alpaca~\cite{cui2023efficient}).
To facilitate Chinese-specific demands, several LLMs trained from scratch exhibit remarkable Chinese proficiency due to multilingual data resources, such as the Baichuan~\cite{Baichuan,yang2023baichuan}, General Language Model (GLM)~\cite{du2021glm,zeng2022glm}, and Qwen~\cite{bai2023qwen} families.
In this work, the Baichuan2-Base series is utilized as the foundation model for our PediatricsGPT, given its comprehensive potential among similar contenders.

\textbf{LLMs in Medical Applications.}
Current LLMs provide unprecedented opportunities to develop resource-efficient and diagnostic-comprehensive intelligent healthcare systems.
Despite universal models~\cite{achiam2023gpt,ChatGPT} equipped with certain internal knowledge regarding biomedicine, they are incompetent in real-world medical applications due to the absence of domain-specific disciplines.
In this context, several efforts~\cite{wang2023huatuo,xiong2023doctorglm,chen2023bianque,li2023chatdoctor} attempt to construct medically tailored LLMs from multiple perspectives.
For instance, ChatDoctor~\cite{li2023chatdoctor} uses patient-doctor conversation data based on LLaMA~\cite{touvron2023llama} to enhance the language model's accuracy in healthcare. DoctorGLM~\cite{xiong2023doctorglm} proves that a healthcare-purpose LLM can be implemented with affordable overhead by fine-tuning ChatGLM-6B~\cite{du2021glm}.
After that, more Chinese medical LLMs~\cite{MedicalGPT,bao2023disc,zhang2023huatuogpt,chen2023huatuogpt2,yang2024zhongjing} are progressively presented to generate doctor-like robust responses, such as HuatuoGPT~\cite{zhang2023huatuogpt}, DISC-MedLLM~\cite{bao2023disc}, and Zhongjing~\cite{yang2024zhongjing}.
Despite advances in general medical knowledge, current models are suboptimal for pressing pediatric applications. In comparison, our sophisticated training procedure and high-quality instruction datasets inject new insights and prospects for developing specialized LLMs with pediatric expertise.

\begin{table}[t]
\setlength{\tabcolsep}{3pt}
\centering
\caption{Statistical information on the proposed dataset. PedCorpus is well extensible and adaptable by incorporating general domain data and as seed instructions to generate specialized corpora (\textit{i.e.,} PedCorpus-CPT and PedCorpus-DFPO). ``KG'' means the Knowledge Graphs.}
\vspace{-2pt}
\resizebox{\linewidth}{!}{%
\begin{tabular}{cccccccc}
\toprule
\multirow{2}{*}{Dataset}       & \multirow{2}{*}{Data Sources}     & \multirow{2}{*}{Department} & \multirow{2}{*}{Number/Size} & \multirow{2}{*}{\begin{tabular}[c]{@{}c@{}}Human \\ Preference\end{tabular}} & \multicolumn{3}{c}{Task Type} \\ \cline{6-8} 
                               &                                   &                             &                              &                                                                              &\rule{0pt}{10pt} MedKQ\&A & EviDiag & TreRecom \\ \midrule
\multirow{5}{*}{PedCorpus}     & Pediatric Textbooks               & Pediatrics                  & 37,284                       & \ding{52}                                                                          & \ding{52}      & --     & \ding{52}      \\
                               & Pediatric Guidelines              & Pediatrics                  & 63,129                       & \ding{52}                                                                          & \ding{52}       & --     & \ding{52}      \\
                               & Pediatric KG                      & Pediatrics                  & 46,320                       & \ding{52}                                                                          & \ding{52}      & --      & \ding{52}      \\
                               & Real Doctor-Patient Conversations & Multiple                    & 46,385                       & \ding{52}                                                                          & --      & \ding{52}     & \ding{52}      \\
                               & Distilled Medical Datasets        & Multiple                    & 107,177                      & --                                                                          & \ding{52}      & \ding{52}      & \ding{52}      \\ \midrule
\multirow{3}{*}{PedCorpus-CPT} & Plain Textbooks, Guidelines, KG      & Multiple                    & \multirow{3}{*}{975.8MB}     & --                                                                          & \ding{52}       & \ding{52}     & \ding{52}      \\
                               &  Filtered Chinese Wikipedia   & Multiple                    &                              & --                                                                          & --      & --      & --      \\
                               & Extended data from PedCorpus                 & Multiple                    &                              & --                                                                          & --      & --     & --      \\ \midrule
PedCorpus-DFPO                  & Pediatrics data from PedCorpus       & Pediatrics                  & 15,556                       & \ding{52}                                                                          & \ding{52}       & \ding{52}     & \ding{52}      \\ \toprule
\end{tabular}
}
\label{tab1}
\vspace{-12pt}
\end{table}

\vspace{-5pt}
\section{Methodology}
\begin{figure}[t]
  \centering
  \includegraphics[width=\linewidth]{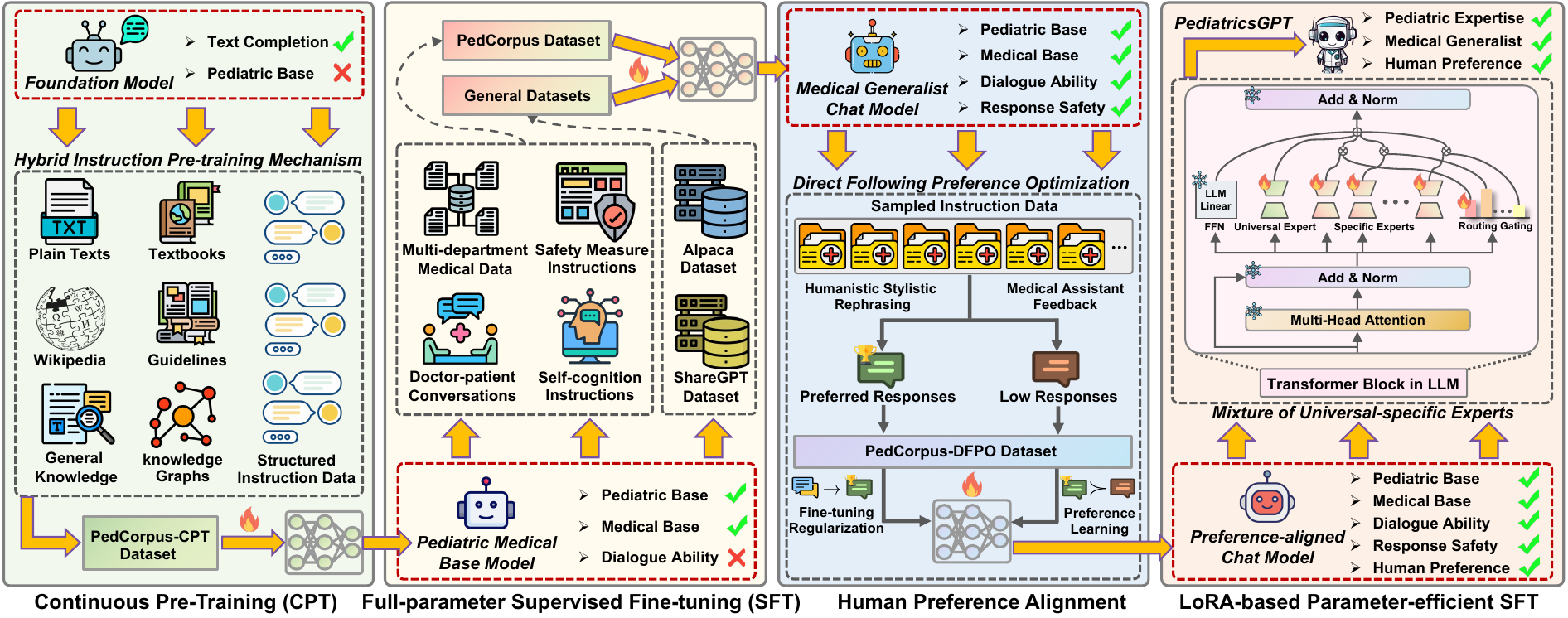}
  \caption{The sequential pipeline for developing PediatricsGPT. We begin by injecting intensive medical and world knowledge into the foundation model through the hybrid instruction mechanism in CPT phase.
  Then, full-parameter SFT is implemented to improve the model's instruction-following capabilities regarding medical generalists. After that, we introduce the direct following preference optimization to control the model behaviour to align with human preference.
  In the parameter-efficient SFT phase, the LoRA-based mixture of universal-specific experts is devised to mitigate conflicts across downstream tasks and competition between pediatric expertise and general mastery.}
  \label{arc}
  \vspace{-0.35cm}
\end{figure}

\vspace{-5pt}
This section describes the proposed PedCorpus dataset and the sequential pipeline for developing PediatricsGPT. Figure~\ref{arc} illustrates the comprehensive method workflow.

\vspace{-3pt}
\subsection{PedCorpus: Multi-task Medical Instruction Dataset}
\vspace{-3pt}
To endow the model with versatile diagnostic proficiency, PedCorpus is constructed through the multi-dimensional corpus across three application-oriented medical tasks, including Knowledge Question-Answer (MedKQ\&A), Evidence-based Diagnosis (EviDiag), and Treatment Recommendation (TreRecom). Table~\ref{tab1} shows the detailed statistical information from different data sources.
We explain the three patterns of PedCorpus construction below.

\textbf{Specialized Pediatric Data.}
Extracting pediatric data from textbooks, guidelines, and knowledge graphs ensures knowledge professionalism.
Specifically, we automatically extract standard medical definitions and descriptions from physical textbooks covering 131 disease types in 11 broad categories.
Over 500 corresponding disease guidelines are collected, including diagnostic protocols and treatment consensus. Additionally, extensive knowledge entities are sampled from ternary instances in the knowledge graphs.
Based on these resources, we introduce a role-playing-driven instruction building rule via GPT-4 API that produces well-organized instructions to enable \textbf{accurate} and \textbf{humanistic} model responses. 
The detailed building procedure is shown in Appendix~\ref{app_sec:1_1}.

\textbf{Real Doctor-patient Conversations.}
To avoid the model collapse dilemma~\cite{shumailov2023curse}, we incorporate authentic doctor-patient dialogues from online treatment platforms and voice transcriptions during medical consultations.
The single-/multi-turn instructions are jointly considered to equip the model with healthcare interrogation and contextual understanding.
Original responses from real doctors are usually terse and noisy, potentially worsening the generation quality~\cite{zhang2023huatuogpt}.
To this end, we craft 100 high-quality examples to guide the advanced language model by the in-context learning to regularize vanilla conversations in the self-instruct pattern~\cite{chen2023phoenix,wang2022self}. This approach ensures \textbf{doctor-like} and \textbf{patient-friendly} model responses.
More regularization details are shown in Appendix~\ref{app_sec:1_2}.

\textbf{Distilled Medical Datasets.}
Integrating general medical knowledge from existing datasets~\cite{li2023huatuo,he2020meddialog,zhang2018multi} is a common practice in previous efforts~\cite{chen2023bianque,xiong2023doctorglm,MedicalGPT,bao2023disc}.
However, we find numerous unclear and incomplete representations in the instruction instances from public benchmarks due to the absence of careful calibration, potentially triggering hallucinated outputs.
Consequently, we manually sample 107,177 knowledge-intensive instructions from three mainstream benchmarks (\textit{i.e.,} Huatuo-26M~\cite{li2023huatuo}, MedDialog~\cite{he2020meddialog}, and CMeKG~\cite{byambasuren2019preliminary}), adhering to the philosophy of quality over quantity~\cite{zhou2024lima}.
After that, a progressive instruction reconstruction rule is proposed to distill the sampled instructions to ensure \textbf{informative} and \textbf{logical} model responses.
The rule process can be found in Appendix~\ref{app_sec:1_3}.

\vspace{-3pt}
\subsection{Hybrid Instruction Pre-training in CPT}
\vspace{-3pt}
Continuous Pre-Training (CPT) is essential in developing domain-specific models~\cite{chen2023huatuogpt2,wen2023chathome,yang2024zhongjing} since it can break the scaling law~\cite{gunasekar2023textbooks} to a certain extent.
For this purpose, we introduce the PedCorpus-CPT dataset to ensure a high-quality pre-training corpus.
From Table~\ref{tab1}, PedCorpus-CPT consists of three-part data components.
(i) We integrate plain texts from vanilla pediatric textbooks, guidelines, and knowledge graphs. 
(ii) The filtered Chinese Wikipedia~\cite{Wikipedia} is also considered to achieve the model's trade-off for medical-general knowledge memory capacity.
(iii) In practice, we observe that CPT leads to catastrophic forgetting of the models at follow-up due to different data distribution and format discrepancies compared to the original pre-training and SFT.
Thus, we introduce a hybrid instruction pre-training mechanism to bridge these discrepancies. 
The core philosophy is to assemble instruction data from PedCorpus with {\tt Input-Output} forms into {\tt Completion} forms, which are then assimilated into plain texts to provide multi-task and complementary information.
This mechanism effectively mitigates inconsistencies between the internal-injected medical knowledge of the foundation model while reinforcing medical domain adaptation.
Moreover, we take PedCorpus as the seed instructions to improve multiple-department corpus density and breadth via knowledge-enhanced prompts.
The prompt template is shown in Appendix~\ref{app_sec:2}. 
 
We pre-train the foundation model to follow the causal language modelling paradigm. Given any input token sequence $\bm{t} = (t_0, t_1, t_2, ...) \in \mathcal{D}_{cpt}$ from the above multi-channel corpus $\mathcal{D}_{cpt}$, the next token $t_i$ is autoregressively predicted by minimizing the negative log-likelihood:
\begin{equation}
\mathcal{L}_{\mathrm{CPT}}(\theta,\mathcal{D}_{cpt}) = \mathbb{E}_{\bm{t}\sim \mathcal{D}_{cpt}}
\left [ -  {\textstyle \sum_{i}^{\left | \bm{t} \right |}}  \mathrm{log}\,p(t_{i} \mid t_0, t_1, ..., t_{i-1};\theta )  \right ],
\end{equation}
where $\theta$ is the model parameter and the input context consists of $t_0, t_1, ..., t_{i-1}$.

\vspace{-3pt}
\subsection{Full-parameter Supervised Fine-tuning}
\vspace{-3pt}

During this phase, we activate the model's ability to follow medical instructions by the Full-parameter Supervised Fine-tuning (FSFT).
The full-parameter pattern enables a fuller invocation of the intensive knowledge in CPT and promotes comprehension and logical reasoning about diverse structured instructions.
The training data at this phase is composed of the following three aspects.
(i) We utilize the multi-department medical data in the PedCorpus dataset to develop the medical generalist.
(ii)
Chinese instruction data (\textit{i.e.,} Alpaca dataset~\cite{peng2023instruction} and ShareGPT~\cite{ShareGPT}) from general domains are selectively integrated to avoid the potential overfitting risk.
(iii)
Providing safety measures is vital for LLM assistants yet overlooked by prior methods~\cite{zhou2024lima}.
In contrast, we write 200 training instructions with some degree of maliciousness, hallucinations, and counterfactuals. Correspondingly, the refusal responses with detailed explanations for disobedience are carefully crafted.
We also include 300 examples related to self-cognition content. These data significantly improve the robustness and security of the model against unfriendly commands.

Given any input instruction $\bm{x} = (x_0, x_1, x_2, ...) \in \mathcal{D}_{fsft}$ and corresponding target response $\bm{y} = (y_0, y_1, y_2, ...) \in \mathcal{D}_{fsft}$ from the above-integrated fine-tuning dataset $\mathcal{D}_{fsft}$, 
the optimization objective can be formulated as follows:
\begin{equation}
\mathcal{L}_{\mathrm{FSFT}}(\theta,\mathcal{D}_{fsft}) = \mathbb{E}_{(\bm{x},\bm{y})\sim \mathcal{D}_{fsft}}
\left [ -  {\textstyle \sum_{i=1}^{\left | \bm{y} \right | }}  \mathrm{log}\,p(y_{i} \mid \bm{x}, y_{<i}; \theta)  \right ].
\end{equation}

\vspace{-3pt}
\subsection{Direct Following Preference Optimization}
\vspace{-3pt}
Aligning human intention preferences facilitates the model to generate harmless responses.
To this end, we introduce PedCorpus-DFPO $\mathcal{D}_{dfpo}$, a preference dataset to guide the model in learning human preference behaviours.
PedCorpus-DFPO contains the input instruction set $\bm{x} = (x_0, x_1, x_2, ...) \in \mathcal{D}_{dfpo}$, which is selectively sampled from vanilla PedCorpus.
On the one hand, we perform a humanistic stylistic rephrasing of the outputs to generate preferred responses $\bm{y}^w = (y^w_0, y^w_1, y^w_2, ...) \in \mathcal{D}_{dfpo}$.
On the other hand, the corresponding low responses $\bm{y}^l = (y^l_0, y^l_1, y^l_2, ...) \in \mathcal{D}_{dfpo}$ are generated from the feedback of a low-capability medical assistant~\cite{wang2023huatuo} to maintain domain consistency.

Despite impressive improvements achieved by RLHF-based approaches~\cite{yang2024zhongjing,zhang2023huatuogpt}, challenges remain due to unstable reward modelling and significant computational costs~\cite{schulman2017proximal,zheng2023secrets}.
Inspired by single-stage preference learning~\cite{rafailov2024direct}, we propose a stable and lightweight method for domain-specific LLMs called Direct Following Preference Optimization (DFPO).
DFPO utilizes variable changes to formulate the preference loss as a policy function that efficiently optimizes the policy with a simple binary cross-entropy objective. 
Meanwhile, our method directly regularizes model behaviour boundaries in an instruction-following paradigm on medical demonstrations of preferred responses,  facilitating robustness and smoothing of the preference learning.

Theoretically, the observed probability of a particular preference pair usually follows the Bradley-Terry model~\cite{bradley1952rank}, and $\bm{y}^{w}$ is preferred over $\bm{y}^{l}$ (denoted $\bm{y}^w \succ \bm{y}^l$):
\begin{equation}
p(\bm{y}^w \succ \bm{y}^l) = \sigma(\gamma (\bm{x}, \bm{y}^w) - \gamma (\bm{x}, \bm{y}^l)),
\end{equation}
where $ \gamma (\bm{x},\bm{y}^{w/l})$ means the parameterized reward function and $\sigma(\cdot)$ is the sigmoid activation.
In this case, the overall optimization objective is expressed as:
\begin{equation}
\mathcal{L}_{\mathrm{DFPO}}(\theta,\mathcal{D}_{dfpo}) = 
-\mathbb{E}_{(\bm{x}, \bm{y}^{w}, \bm{y}^{l})\sim \mathcal{D}_{dfpo}}
\left [ 
\mathrm{log}\,\sigma (\beta\, \mathrm{log} \frac{\pi_{\theta}(\bm{y}^{w}\mid \bm{x}) }{\pi_{r}(\bm{y}^{w}\mid \bm{x})} - \beta\, \mathrm{log} \frac{\pi_{\theta}(\bm{y}^{l}\mid \bm{x}) }{\pi_{r}(\bm{y}^{l}\mid \bm{x})})
\right ] + \mu \Phi(\bm{x}, \bm{y}^w),
\end{equation}
where $\pi_{\theta}$ and $\pi_{r}$ are the desired optimal policy and the reference policy, respectively.
$\beta$ is the control parameter reflecting the deviation from the basic $\pi_{r}$.
For the fine-tuning regularization term $\Phi(\bm{x}, \bm{y}^w)$ with the scaling coefficient $\mu$, the implementation process is equivalent to maximizing the log probability $p(\bm{y}^w \mid \bm{x})$ regarding the preferred responses $\bm{y}^w$ given the input instructions $\bm{x}$:
\begin{equation}
\Phi(\bm{x}, \bm{y}^w) = \mathbb{E}_{(\bm{x},\bm{y}^w)\sim \mathcal{D}_{dfpo}}
\left [ -  {\textstyle \sum_{i=1}^{\left | \bm{y}^w \right | }}  \mathrm{log}\,p(y_{i}^w \mid \bm{x}, y_{<i}^w; \theta)  \right ].
\end{equation}

\subsection{Mixture of Universal-specific Experts in Parameter-efficient SFT}
This phase aims to reinforce the model performance for various pediatric applications through the LoRA-based Parameter-efficient SFT (PSFT). The used dataset $\mathcal{D}_{psft}$ is derived from the pediatric department in PedCorpus and partial general medical/world data.
In practice, we observe that competition across different pediatric tasks and the conflicts between medical generalization and specialized knowledge deteriorate instruction-following abilities.
Accordingly, we propose a mixture of universal-specific experts strategy to address these challenges.
Formally, LoRA adapters~\cite{hu2021lora} act as experts to replace the linear layers in the Feed-Forward Neural (FFN) networks of LLMs, providing trainable parameters.
Several specific experts $\left \{ E_j^{s} \right \}^{T}_{j=1}$ are assigned adaptive activations to master distinct pediatric expertise through soft routing. The routing gating is defined as follows:
\begin{equation}
G(\bm{x}) = \mathrm{Softmax} (\bm{x}\bm{W}_{g}+ \mathcal{S}(\varphi (\bm{x}\bm{W}_{n})).
\end{equation}
$\bm{W}_{g}$ and $\bm{W}_{n}$ are the learnable weights. $\mathcal{S}(\varphi (\bm{x}\bm{W}_{n})$ is the noise term for regularizing the expert utilization balance, where $\mathcal{S}(\cdot)$ and $\varphi(\cdot)$ represent the Standard Normal distribution sampling and Softplus function, respectively.
Moreover, we consistently activate a universal expert $E^{u}$ across all training data to prevent general knowledge forgetting and mitigate competency conflict.
The parameterized output $\bm{z}$ of all the experts in the forward process can be mathematized as follows:
\begin{equation}
\bm{z} = \frac{\alpha}{r}({\sum_{j=1}^{T}} G(\bm{x})_j E_j^{s}(\bm{x}) + E^{u}(\bm{x})),
\end{equation}
where $r$ is the rank value and $\alpha$ is a hyper-parameter for approximating the learning rate.

\vspace{-6pt}
\section{Experiments}

\begin{table}[t]
\setlength{\tabcolsep}{2pt}
\centering
\caption{Comparison results of different models on three pediatric medical benchmarks. In each benchmark, the best results are marked in \textbf{bold}, and the second-best results are marked \underline{underlined}.}
\vspace{-2pt}
\resizebox{\linewidth}{!}{%
\begin{tabular}{clccccccccccc}
\toprule
\multicolumn{2}{c}{Benchmark}                 & Model             & ROUGE-1        & ROUGE-2        & ROUGE-L        & BLEU-1         & BLEU-2         & BLEU-3         & BLEU-4         & GLEU           & Distinct-1     & Distinct-2     \\ \midrule
\multicolumn{2}{c}{\multirow{9}{*}{MedKQ\&A}}   & Baichuan2-7B      & 40.88          & 19.44          & 21.50           & 26.77          & 20.00             & 17.30           & 14.86          & 24.88          & 20.14          & 39.95          \\
\multicolumn{2}{c}{}                          & Baichuan2-13B     & 46.96          & 22.85          & 22.54          & 29.02          & 25.62          & 22.63          & 19.31          & 27.97          & 21.45          & 42.53          \\
\multicolumn{2}{c}{}                          & HuatuoGPT        & 48.52          & 23.44          & 25.13          & 43.00             & 41.25          & 36.31          & 29.82          & 34.60           & 20.42          & 41.27          \\
\multicolumn{2}{c}{}                          & DISC-MedLLM              & 53.83          & 25.98          & 27.71          & 47.91          & 44.57          & 37.65          & 30.07          & 37.11          & \underline{26.63} & \underline{51.98} \\
\multicolumn{2}{c}{}                          & Zhongjing         & 53.97          & 26.03          & 29.56          & 51.11          & 45.04          & 39.13          & 33.59          & 42.61          & \textbf{26.75} & \textbf{52.66} \\
\multicolumn{2}{c}{}                          & HuatuoGPT-II       & 55.27          & 26.59          & 27.95          & 59.07          & 51.49          & 45.38          & 38.70           & 39.18          & 20.97          & 41.34          \\
\multicolumn{2}{c}{}                          & Meditron-7B & 55.63 & 26.19 & 30.37 & 58.43 & 53.45 & 56.07 & 38.77 & 42.23 & 22.34 & 45.17 \\
\multicolumn{2}{c}{}                          & Llama3.1-8B &  53.18 &  24.74 &  28.26 &  45.07 &  42.45 &  36.57 &  29.73 &  35.63 &  22.74 &  46.52 \\
\multicolumn{2}{c}{}                          & ChatGPT               & 56.92          & 27.87          & 29.05          & 61.58 & 54.37          & 47.97          & 40.77          & 45.15          & 20.76          & 40.19          \\
\multicolumn{2}{c}{}                          & GPT-4 & \underline{58.79} & \underline{33.56} & \underline{32.15} & \textbf{62.53} & \underline{59.14} & 55.26 & 52.39 & 53.72 & 21.79 & 43.26 \\
\multicolumn{2}{c}{}                          &  PediatricsGPT-7B  & 58.08 & 31.78 & 31.11 & 59.41          & 56.88 & \underline{57.47} & \underline{55.34} & \underline{54.41} & 24.33          & 47.41          \\
\multicolumn{2}{c}{}                          & PediatricsGPT-13B & \textbf{60.85} & \textbf{36.56} & \textbf{35.64} & \underline{61.65} & \textbf{63.17} & \textbf{58.96} & \textbf{59.34} & \textbf{57.22} & 24.24          & 46.23          \\ \midrule
\multicolumn{2}{c}{\multirow{9}{*}{EviDiag}}  & Baichuan2-7B      & 26.81          & 7.75           & 11.22          & 15.18          & 11.51          & 9.19           & 6.72           & 13.44          & 23.65          & 46.93          \\
\multicolumn{2}{c}{}                          & Baichuan2-13B     & 39.14          & 12.06          & 12.44          & 47.65          & 36.02          & 28.82          & 21.19          & 28.28          & 25.45          & 50.43          \\
\multicolumn{2}{c}{}                          & HuatuoGPT        & 35.12          & 10.77          & 15.04          & 46.22          & 33.10           & 25.44          & 21.22          & 25.44          & 22.30           & 45.73          \\
\multicolumn{2}{c}{}                          & DISC-MedLLM              & 33.55          & 11.67          & 15.32          & 15.91          & 12.46          & 10.27          & 7.96           & 16.77          & \textbf{35.89} & \textbf{69.36} \\
\multicolumn{2}{c}{}                          & Zhongjing         & 40.92          & 14.26          & 17.41          & 48.64          & 37.52          & 30.17          & 22.44          & 27.03          & \underline{33.40}  & \underline{65.89} \\
\multicolumn{2}{c}{}                          & HuatuoGPT-II       & 39.52          & 12.14          & 16.38          & 49.58          & 37.62          & 30.66          & 23.34          & 28.98          & 21.97          & 43.62          \\
\multicolumn{2}{c}{}                          & Meditron-7B & 42.63 & 15.12 & 18.94 & 52.36 & 39.24 & 37.78 & 27.15 & 31.25 & 22.07 & 45.13 \\
\multicolumn{2}{c}{}                          & Llama3.1-8B &  37.25 &  13.07 &  16.23 &  44.54 &  32.29 &  23.72 &  20.12 &  23.57 &  24.43 &  45.67 \\
\multicolumn{2}{c}{}                          & ChatGPT               & 40.88          & 13.42          & 16.97          & 48.84          & 37.69          & 30.55          & 23.17          & 29.02          & 23.49          & 46.54          \\ 
\multicolumn{2}{c}{}                          & GPT-4 & \textbf{48.48} & \underline{16.74} & \underline{21.51} & \underline{57.59} & \underline{44.78} & \underline{37.94} & \underline{30.56} & \underline{36.79} & 25.69 & 50.13 \\
\multicolumn{2}{c}{}                          &  PediatricsGPT-7B  & 45.83 & 16.60  & 19.91 & 54.37 & 41.99 & 37.59 & 29.03 & 33.42 & 23.49          & 46.61          \\
\multicolumn{2}{c}{}                          & PediatricsGPT-13B & \underline{47.32} & \textbf{17.63} & \textbf{21.87} & \textbf{58.21} & \textbf{45.72} & \textbf{39.74} & \textbf{31.25} & \textbf{37.15} & 23.34          & 46.34          \\ \midrule
\multicolumn{2}{c}{\multirow{9}{*}{TreRecom}} & Baichuan2-7B      & 48.39          & 23.07          & 26.35          & 47.94          & 40.91          & 35.54          & 29.69          & 35.06          & 21.90           & 43.57          \\
\multicolumn{2}{c}{}                          & Baichuan2-13B     & 48.87          & 23.41          & 26.42          & 49.96          & 46.24          & 42.84          & 35.04          & 35.63          & 22.36          & 45.12          \\
\multicolumn{2}{c}{}                          & HuatuoGPT        & 53.48          & 25.41          & 27.08          & 58.14          & 49.64          & 42.93          & 35.16          & 41.63          & 23.26          & 46.21          \\
\multicolumn{2}{c}{}                          & DISC-MedLLM              & 52.77          & 24.26          & 28.89          & 58.73          & 50.05          & 42.96          & 35.59          & 42.44          & 24.30           & 51.95          \\
\multicolumn{2}{c}{}                          & Zhongjing         & 54.92          & 26.63          & 29.68          & 60.12          & 53.31          & 44.25          & 38.76          & 40.38          & 26.18          & 53.94          \\
\multicolumn{2}{c}{}                          & HuatuoGPT-II       & 58.44          & 30.47          & 32.02          & 59.91          & 54.26          & 45.73          & 38.92           & 42.28          & 28.88          & 57.15 \\
\multicolumn{2}{c}{}                          & Meditron-7B & 58.56 & 32.25 & 33.37 & 60.47 & 55.36 & 48.73 & 42.18 & 46.73 & 28.51 & \underline{57.45} \\
\multicolumn{2}{c}{}                          & Llama3.1-8B &  52.45 &  24.98 &  26.14 &  57.56 &  48.11 &  41.67 &  24.03 &  40.67 &  22.73 &  45.13 \\
\multicolumn{2}{c}{}                          & ChatGPT           & 59.59 & 33.34 & 35.79 & 62.81 & 55.79 & 49.85 & 43.29 & 47.59 & \underline{31.09} & 56.87          \\ 
\multicolumn{2}{c}{}                          & GPT-4 & \underline{61.94} & \underline{37.27} & \underline{36.73} & \underline{63.23} & \underline{56.24} & \underline{50.58} & \underline{44.07} & \textbf{55.26} & 30.27 & 56.48 \\
\multicolumn{2}{c}{}                          &  PediatricsGPT-7B  & 56.92          & 29.13          & 31.26          & 61.36          & 55.34          & 46.44          & 40.61          & 44.65          & 26.06          & 52.77          \\
\multicolumn{2}{c}{}                          & PediatricsGPT-13B & \textbf{62.83} & \textbf{39.32} & \textbf{40.82} & \textbf{63.56} & \textbf{56.68} & \textbf{50.80}  & \textbf{44.31} & \underline{54.65} & \textbf{31.94} & \textbf{57.56} \\ \bottomrule
\end{tabular}
}
\label{tab2}
\vspace{-6pt}
\end{table}

\vspace{-3pt}
\subsection{Datasets and Implementation Details}
\vspace{-3pt}
Extensive experiments are conducted on three application-oriented benchmarks to assess the model's pediatric medical abilities, including Knowledge Question-Answer (\textbf{MedKQ\&A}), Evidence-based Diagnosis (\textbf{EviDiag}), and Treatment Recommendation (\textbf{TreRecom}).
Each benchmark contains 300 held-out samples to reject data leakage during training.
In addition, we select two publicly available Chinese medical benchmarks to validate the model's generalizability in general healthcare.
Specifically, we sample 50 challenging instances of diagnostic queries from each department from the \textbf{webMedQA}~\cite{he2019applying} and \textbf{CMD}~\cite{CMD} benchmarks, respectively, leading to testing sets with 300 samples.

Our PediatricsGPT is developed upon the Baichuan2-Base~\cite{yang2023baichuan} models in two versions with 7 and 13 billion parameters.
The model training is accomplished through the PyTorch platform with Accelerate and DeepSpeed packages using eight Nvidia A800 GPUs.
The ZeRO strategy~\cite{rajbhandari2020zero} is employed to alleviate the memory overhead during full parameter training.
The AdamW optimizer~\cite{loshchilov2017decoupled} is adopted for network optimization, and the bf16 data accuracy is chosen. More detailed hyper-parameter configurations for different stages are shown in Appendix~\ref{app_sec:3}.

\vspace{-3pt}
\subsection{Model Zoo}
\vspace{-3pt}
We compare a series of LLMs for comprehensive evaluations. Concretely, \textbf{Baichuan2-7B/13B (Chat)} models~\cite{yang2023baichuan} are trained on 2.6 trillion tokens as the baselines, which have excellent abilities in different domains. 
\textbf{Meditron-7B}~\cite{chen2023meditron} is a 7 billion parameters model adapted to the medical domain from Llama2-7B through continued pre-training on a comprehensively curated medical corpus.
\textbf{Llama3.1-8B}~\cite{dubey2024llama} is a robust multilingual large language model through systematic training.
For reproducible Chinese medical works, \textbf{DISC-MedLLM (13B)}~\cite{bao2023disc} is fine-tuned through reconstructed medical dialogues and behavioural preference instructions.
\textbf{HuatuoGPT (13B)}~\cite{zhang2023huatuogpt} performs SFT based on mixed instruction data and introduce human feedback in RLHF.
\textbf{HuatuoGPT-II (13B)}~\cite{chen2023huatuogpt2} enhances the medical-specific domain adaptation of LLMs through one-stage unified training.
\textbf{Zhongjing (13B)}~\cite{yang2024zhongjing} implements a complete pipeline based on Ziya-LLaMA-13B to enhance the model's multi-turn medical conversation abilities.
\textbf{ChatGPT}~\cite{ChatGPT} and \textbf{GPT-4}~\cite{achiam2023gpt} have impressive performance in general medical fields as closed-source models developed by OpenAI.

\vspace{-3pt}
\subsection{Comparison with State-of-the-art Methods}
\vspace{-3pt}

 \begin{figure}[t]
  \centering
  \includegraphics[width=\linewidth]{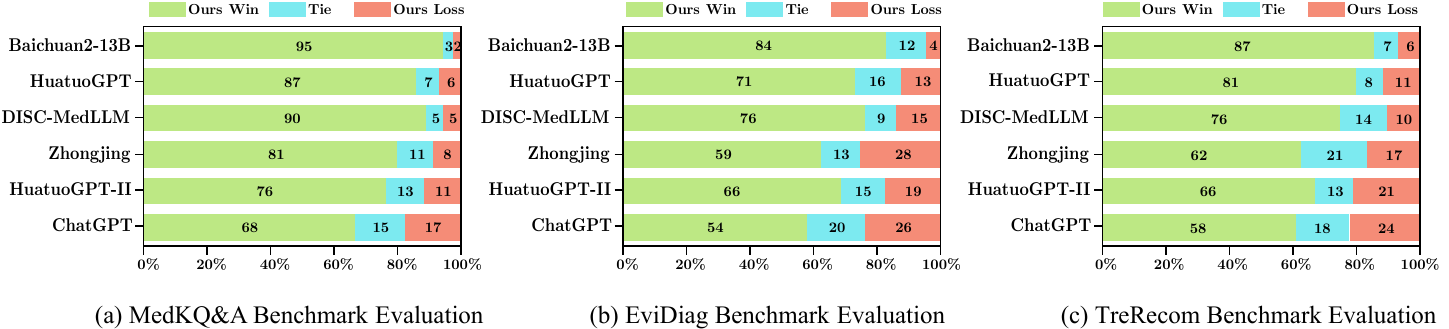}
  \caption{Response comparisons of PediatricsGPT-13B with other baselines via GPT-4 evaluation.}
  \label{gpt_eval}
\end{figure}

 \begin{figure}[t]
  \centering
  \includegraphics[width=\linewidth]{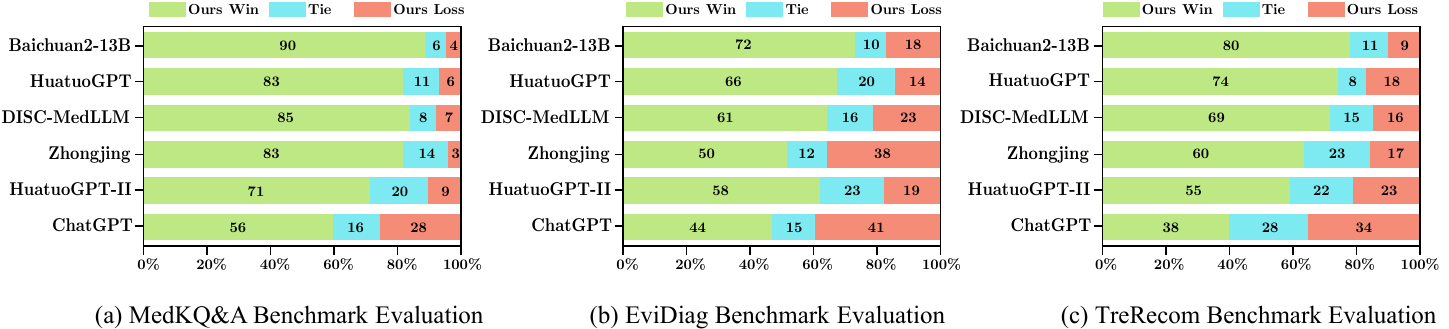}
  \caption{Response comparisons of PediatricsGPT-13B with other baselines via Doctor evaluation.}
  \label{doctor_eval}
\end{figure}

\textbf{Metrics-based Evaluation.}
In Table~\ref{tab2}, we present the comparison results of different models on three pediatric benchmarks through multifaceted metrics, including ROUGE-1/2/L, BLEU-1/2/3/4, GLEU, and Distinct-1/2.
The key observations are listed below.
(i) PediatricsGPT-13B significantly outperforms the baselines and SOTA medical models on the vast majority of metrics across all benchmarks, demonstrating excellent pediatric expertise.
(ii) Our 7B version also achieves competitive results compared to the 13B models. For instance, PediatricsGPT-7B yields absolute improvements of 3.53\% and 4.44\% on metrics ROUGE-L and GLEU in the EviDiag task compared to HuatuoGPT-II, respectively, generating more accurate and informative content.
(iii) By contrast to Zhongjing and HuatuoGPT-II with massive training corpora, our method confirms that the training data quality outweighs quantity for performance gains.
(iv) The worst results at baselines emphasize that target-oriented fine-tuning is an effective strategy for improving domain-specific abilities.

\textbf{Automated GPT-4 Evaluation.}
Measuring model performance from multiple aspects is essential in the pediatric medical domain. 
To this end, we consider four dimensions to holistically assess response quality, including \textit{usefulness}, \textit{correctness}, \textit{consistency}, and \textit{smoothness}.
Advanced GPT-4~\cite{achiam2023gpt} is prompted to select the winning response between pairwise models based on these dimensions.
The dimension explanations and the prompt template for GPT-4 can be found in Appendix~\ref{app_sec:4}.
(i) As Figure~\ref{gpt_eval} shows, PediatricsGPT-13B wins all LLMs by large margins in the MedKQ\&A task, implying the necessity of implementing the knowledge-intensive CPT.
(ii) The favourable win rates on the TreRecom and EviDiag tasks compared to medical LLMs show the superiority of our model in both single-turn treatment recommendations and multi-turn medical diagnostics. 
For example, our model beats Zhongjing via the 59\% win rate on the EviDiag, which specializes in multi-round consultations.

\textbf{Manual Doctor Evaluation.}
Doctor approval of LLM assistants is a vital step toward realistic applications. 
We invite three doctors (each paid \$300) to determine the winner of pairwise models by the majority voting rule. The evaluation requires simultaneous consideration of the responses' \textit{professionalism}, \textit{factuality}, and \textit{safety}.
(i) Excluding ChatGPT, the dominance of our model in Figure~\ref{doctor_eval} shows the effectiveness of considering safety measure data while incorporating specialized pediatric knowledge.
(ii) The proposed direct following preference optimization makes PediatricsGPT-13B more favoured by human preferences compared to other behavioural alignment efforts~\cite{bao2023disc,yang2024zhongjing,zhang2023huatuogpt}.
(iii) The competitive performance of ChatGPT when human judgments indicate that the scaling law still holds, stemming from the high agreement between its behaviours and human intentions.

 \begin{figure}[t]
 \vspace{-5pt}
  \centering
  \includegraphics[width=\linewidth]{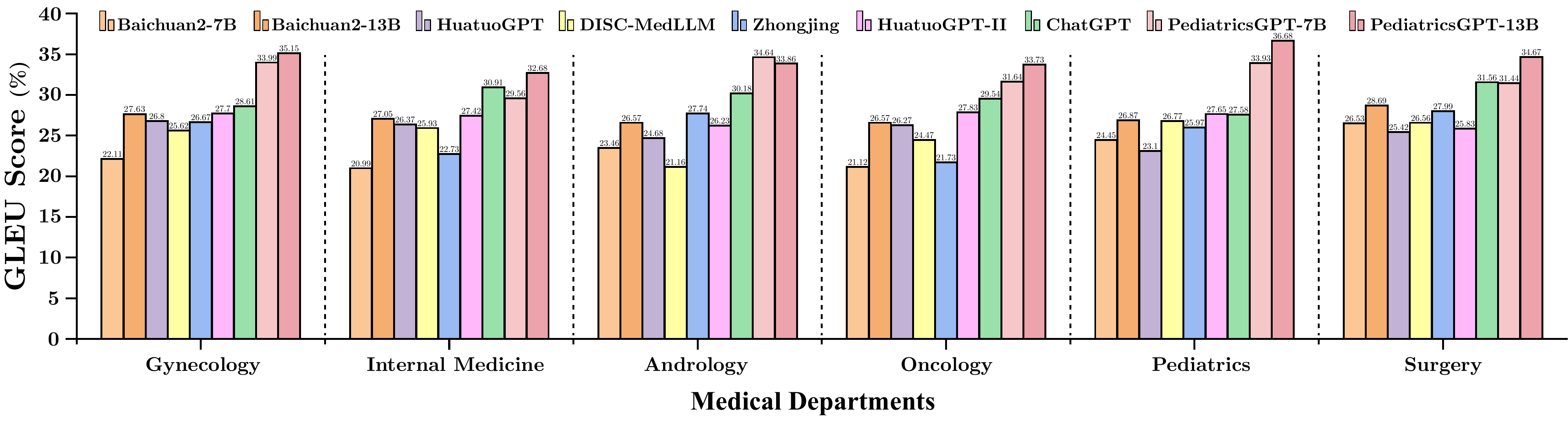}
  \caption{Comparison results of different models on the CMD benchmark.}
  \label{cmd_eval}
\end{figure}

 \begin{figure}[t]
  \centering
  \includegraphics[width=\linewidth]{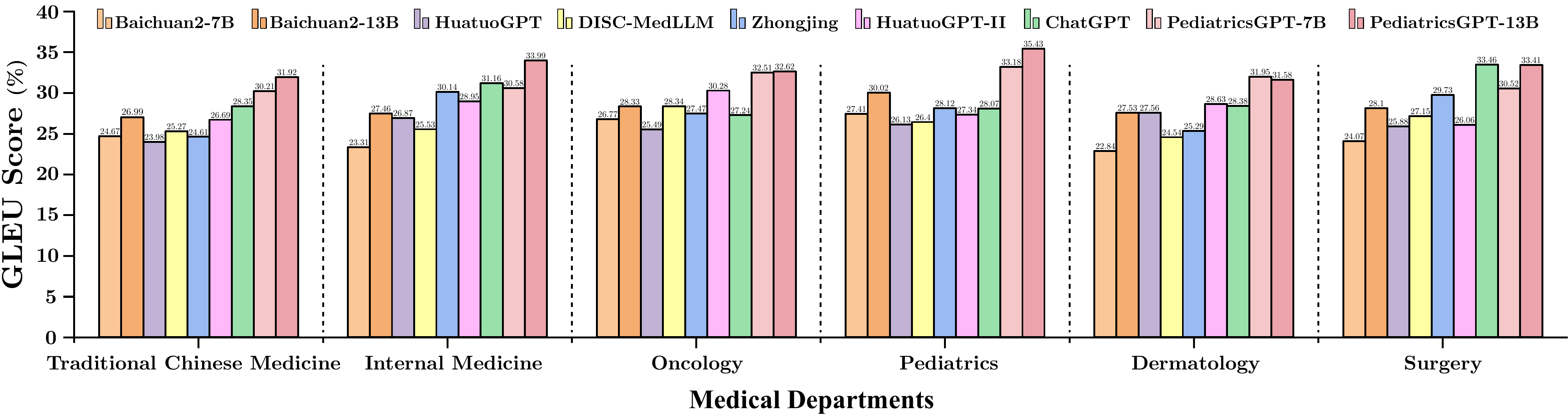}
  \vspace{-5pt}
  \caption{Comparison results of different models on the webMedQA benchmark.}
  \label{web_eval}
  \vspace{-0.2cm}
\end{figure}

\textbf{Generalization Ability Evaluation.}
We show the GLEU metric-based scores of different models on the Chinese medical benchmarks in Figure~\ref{cmd_eval} for CMD and Figure~\ref{web_eval} for webMedQA. (i) PediatricsGPT-13B achieves impressive results across diverse medical departments (including pediatrics), exhibiting medical generalist and pediatric competency mastery.
(ii) The 7B counterpart similarly outperforms most 13B Chinese medical LLMs and exceeds ChatGPT in some departments. For instance, PediatricsGPT-7B brings relative gains of 18.8\% and 7.1\% compared to ChatGPT in the Gynecology and Oncology tasks on the CMD benchmark.
These findings confirm the robust generalization of our model and its ability to capture the multifaceted medical dialogue distributions.

\begin{table}[t]
\setlength{\tabcolsep}{3pt}
\centering
\caption{Ablation study results on five medical benchmarks. ``w/'' and ``w/o'' are short for with and without, respectively. ``MUE'' means the Mixture of Universal-specific Experts strategy.}
\vspace{-2pt}
\resizebox{\linewidth}{!}{%
\begin{tabular}{ccccccccccc}
\toprule
\multirow{2}{*}{Components}                      & \multicolumn{2}{c}{MedKQ\&A} & \multicolumn{2}{c}{EviDiag} & \multicolumn{2}{c}{TreRecom} & \multicolumn{2}{c}{CMD}   & \multicolumn{2}{c}{webMedQA} \\ \cline{2-11} 
                                            & \rule{0pt}{10pt}GPT-4          & Doctor      & GPT-4          & Doctor       & GPT-4           & Doctor       & GPT-4         & Doctor      & GPT-4           & Doctor       \\ \midrule
\textbf{Full Model}                                  & \textbf{68}\%  & \textbf{56}\% & \textbf{54}\%  & \textbf{44}\%  & \textbf{58}\%   & \textbf{38}\%  & \textbf{46}\% & \textbf{40}\% & \textbf{53}\%   & \textbf{45}\%  \\ \midrule
\multicolumn{11}{c}{Importance of Continuous Pre-training}                                                                                                                                       \\ \midrule
w/o Continuous Pre-training                  & 61\%           & 50\%           & 46\%            & 39\%            & 51\%             & 31\%            & 39\%           & 33\%           & 47\%             & 38\%            \\
w/o Hybrid Instruction Pre-training          & 67\%            & 54\%           & 52\%            & 43\%            & 57\%             & 36\%            & 44\%           & 38\%          & 52\%             & 43\%            \\ \midrule
\multicolumn{11}{c}{Necessity of Supervised Fine-tuning}                                                                                                                                         \\ \midrule
w/o Full-parameter SFT                       & 65\%            & 53\%           & 50\%           & 42\%            & 55\%             & 36\%            & 42\%           & 36\%           & 49\%             & 41\%            \\
w/o Parameter-efficient SFT                  & 63\%            & 51\%           & 49\%            & 40\%            & 53\%             & 34\%            & 45\%           & 39\%           & 51\%             & 43\%            \\
w/o MUE Strategy                                      & 67\%   & \textbf{57}\% & 52\%            & 43\%            & 56\%             & 36\%            & 43\%           & 37\%           & 50\%             & 42\%            \\
w/o Universal Expert                         & 67\%            & \textbf{56}\% & 53\%            & \textbf{44}\%  & 57\%             & 37\%            & 44\%           & 38\%           & 51\%             & 42\%            \\ \midrule
\multicolumn{11}{c}{Effectiveness of Preference Alignment}                                                                                                                                       \\ \midrule
w/o DFPO & 67\%            & 53\%           & 52\%            & 41\%            & 57\%             & 36\%            & 45\%           & 38\%          & 51\%             & 41\%            \\
w/ Vanilla DPO            & 66\%            & 55\%           & 53\%            & 42\%            & 57\%             & 36\%            & 44\%           & 38\%           & 52\%             & 42\%            \\
w/  RLHF                                     & 67\%            & 55\%           & \textbf{54}\%  & 43\%            & 57\%            & 37\%            & 45\%           & 39\%           & 52\%             & 44\%            \\ \bottomrule
\end{tabular}
}
\label{tab3}
\end{table}

\subsection{Ablation Studies}
\vspace{-3pt}
We perform thorough ablation studies on five medical benchmarks to investigate the effects of different modelling components. Following~\cite{zhang2023huatuogpt}, we compare the responses from each of the proposed model variants with ChatGPT, and then calculate the win rate (\%) of our model in pairwise responses by GPT-4 and doctor evaluations. Table~\ref{tab3} shows the following observations.

\textbf{Importance of Continuous Pre-training.}
Firstly, we remove the complete continuous pre-training phase to observe performance variations. (i) The significantly deteriorated win rates reveal that injecting specialized knowledge into medical LLMs through rich corpora is indispensable. (ii) Meanwhile, our hybrid instruction pre-training mechanism provides valuable gains to the model.

\begin{wrapfigure}{r}{0.6\textwidth}
\vspace{-12pt}
  \centering
  \setlength{\abovecaptionskip}{0pt}
  \setlength{\belowcaptionskip}{-8pt}
  \includegraphics[width=0.6\textwidth]{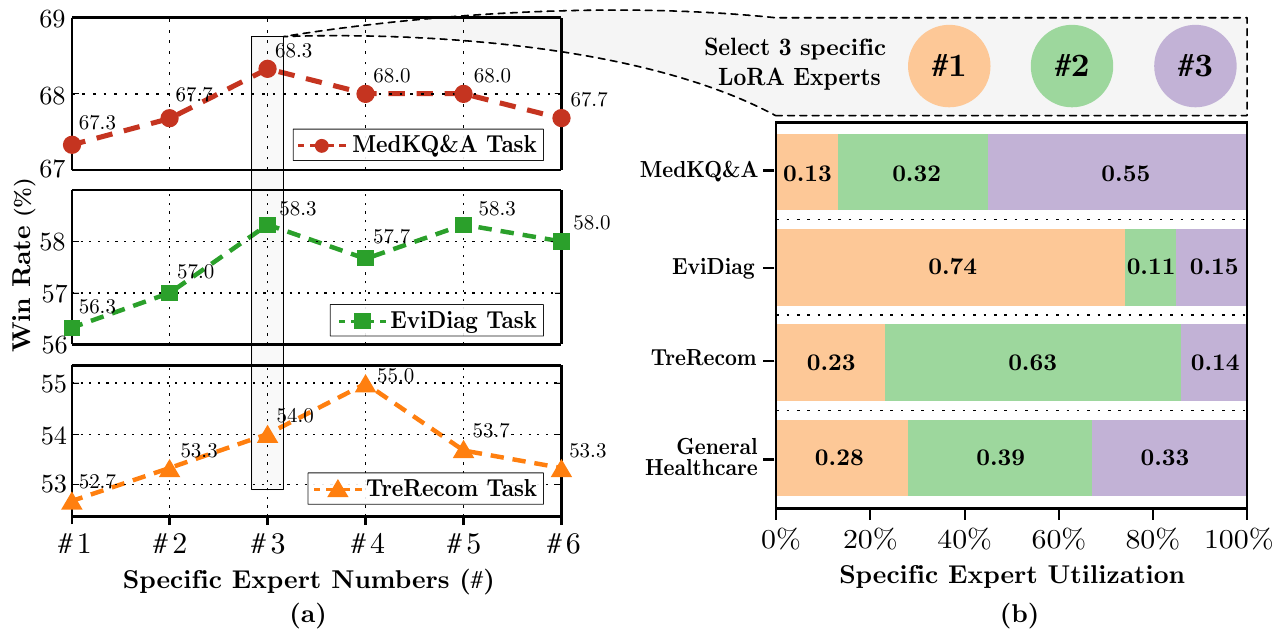}
  \caption{(a) and (b) show the effect of specific expert numbers on model performance and specific expert utilization in different task data, respectively.}
  \label{vis}
\end{wrapfigure}

\textbf{Necessity of Supervised Fine-tuning.}
(i) We observe consistent performance gaps when removing the Full-parameter SFT (FSFT) and Parameter-efficient SFT (PSFT) phases, respectively. This makes sense since SFTs are necessary to activate the model's healthcare instruction-following capabilities.
(ii) Moreover, PSFT is more critical for three pediatric applications because it facilitates pediatric-related knowledge accumulation, while FSFT focuses on consolidating general medical semantic representations.
(iii) Then, we replace the proposed Mixture of Universal-specific Experts (MUE) version with the vanilla single LoRA.
The reduced performance on pediatric EviDiag and TreRecom benchmarks verifies that it is essential to introduce multiple LoRAs that act as specific experts on different tasks. 
A reasonable explanation is that the single-LoRA model suffers from the task competition between learning the knowledge question-answer and mastering the diagnostic recommendation abilities.
(iv) Furthermore, we find that the universal LoRA expert significantly improves the results on the general medical benchmarks (\textit{i.e.}, CMD and webMedQA), proving that it mitigates the competency conflict between general medical and pediatric knowledge.

\textbf{Effectiveness of Preference Alignment.}
(i) When the Direct Following Preference Optimization (DFPO) phase is removed, the model exhibits significant performance drops in doctor evaluations compared to the full version. This observation proves that DFPO effectively helps the model to align human preferences, reducing harmful content while generating doctor-like output.
(ii) As two candidates, the vanilla DPO and RLHF methods are inferior to the proposed DFPO, suggesting that our strategy can more safely control model behaviour, leading to more favoured humanistic responses.

\vspace{-5pt}
\subsection{Qualitative Analysis of LoRA Experts}
\vspace{-5pt}
\textbf{Effect of Specific Expert Numbers.}
As a complement to the ablation of LoRA experts, Figure~\ref{vis}(a) explores the gain effects of varying the number of specific experts while maintaining the universal expert. (i) Noticeably, our MCE strategy with three specific experts achieves a reasonable performance trade-off across the three tasks with only 0.95\% trainable parameters. (ii) Conversely, excessively introducing LoRA experts does not result in appreciable gains but increases the training overhead.

\textbf{Analysis of Expert Utilization.}
To confirm the duties of specific LoRA experts in the routing process, we visualize the normalized weights assigned by the routing gating when encountering data from different downstream tasks.
CMD and webMedQA data are merged to compose general healthcare data.
From Figure~\ref{vis}(b), (i) Experts 2 and 3 are emphatically activated on the TreRecom and MedKQ\&A tasks, respectively, implying their focal ability to tackle medical knowledge interpretations and treatment recommendations.
(ii) In contrast, Expert 1 is more proficient at learning multi-turn diagnosis semantics in the EviDiag task, which is different from the other tasks of instruction content.
(iii) Additionally, there is no clear difference in the specific expert utilization on general healthcare, implying that the general task is handled by the consistently universal expert.
The above observations demonstrate the effectiveness and necessity of the proposed MCE strategy.

\subsection{Visualization Analysis of Model Responses}
To intuitively compare the output quality of medical LLMs, we show the responses of different models for each of the three types of medical inquiries in 
Figures~\ref{tab_medqa}\&\ref{tab_diag}\&\ref{tab_recom} from Appendix~\ref{app_sec:5}.
From the results, Zhongjing offers insufficient information due to limited output content. Although HuatuoGPT-II gives well-organized responses, it lacks accuracy and informativeness. In comparison, our model can provide more specialized and detailed medical knowledge and diagnostic guidance in extended response contexts,  confirming its application potential in diverse healthcare services.

\vspace{-5pt}
\section{Conclusion and Discussion}
\vspace{-5pt}
This paper presents PediatricsGPT, a Chinese medical LLM assistant with medical generalist and pediatric expertise capabilities. Based on the well-designed PedCorpus dataset, PediatricsGPT undergoes a systematic and robust procedure ranging from continuous pre-training and supervised fine-tuning to human preference optimization, leading to competence in different pediatric and general healthcare service scenarios.
Extensive experimental results under multi-dimensional evaluation patterns demonstrate that our model outperforms currently available Chinese medical LLMs, providing a potential solution for promoting reliable and intelligent interactive diagnosis and treatment.

\textbf{Broader Impacts.}
(i) Our model has made meaningful contributions to pediatric medicine by integrating extensive medical data and emerging research. This integration facilitates more accurate and expedited diagnosis of complex pediatric conditions and aids in predicting treatment outcomes, enabling highly personalized and effective treatment strategies for young patients.
(ii) The proposed PediatricsGPT provides crucial decision support for medical professionals, giving evidence-based recommendations and specialized medical insights.  Additionally, it democratizes access to expert medical suggestions and accurate medical knowledge, empowering parents and caregivers with accurate health information, which is especially crucial in underserved areas.
(iii) The training pipeline of PediatricsGPT showcases exemplary generalizability, designed to be applicable across various medical and non-medical domains. This adaptability broadens the model’s applicability and pioneers the development of future AI solutions in healthcare and other fields.

\textbf{Limitations.}
(i) When deployed online, the proposed PediatricsGPT model, like other Large Language Models (LLMs), faces significant security risks, particularly from attacks aimed at manipulating its outputs. These attacks can be strategically designed to exploit the model's response mechanisms, allowing attackers to induce the model to generate unsafe, biased, or otherwise inappropriate content. 
(ii) Currently, our PediatricsGPT model does not support all languages. This linguistic barrier can prevent the model from reaching a global audience, particularly in diverse linguistic landscapes where localized medical information is crucial. 

\textbf{Ethical Issues.}
We fully recognize the critical importance of privacy and data protection. All data used has been meticulously de-identified, with all sensitive information removed, and this process has been verified by the partnering medical institutions.
For the public databases, we strictly follow specific license agreements for use and adaptation. For the constructed corpus, we underwent an internal ethical review by the ethical review board of the partnering medical institutions with license and approval. We will release relevant resources to the extent that they are controlled and permitted.

We provide more discussions of the future work in Appendix \ref{app_sec:8}.

\textbf{Acknowledgment.} This work is supported in part by the National Key R\&D Program of China under Grant 2021ZD0113502, in part by the Shanghai Municipal Science and Technology Major Project under Grant 2021SHZDZX0103.

\bibliographystyle{plain}
\bibliography{nips}

\newpage
\appendix

\section{Implementation Details of PedCorpus Construction}
\subsection{Role-playing-driven Instruction Building Rule}
\label{app_sec:1_1}
\begin{figure}[h]
  \centering
  \includegraphics[width=\linewidth]{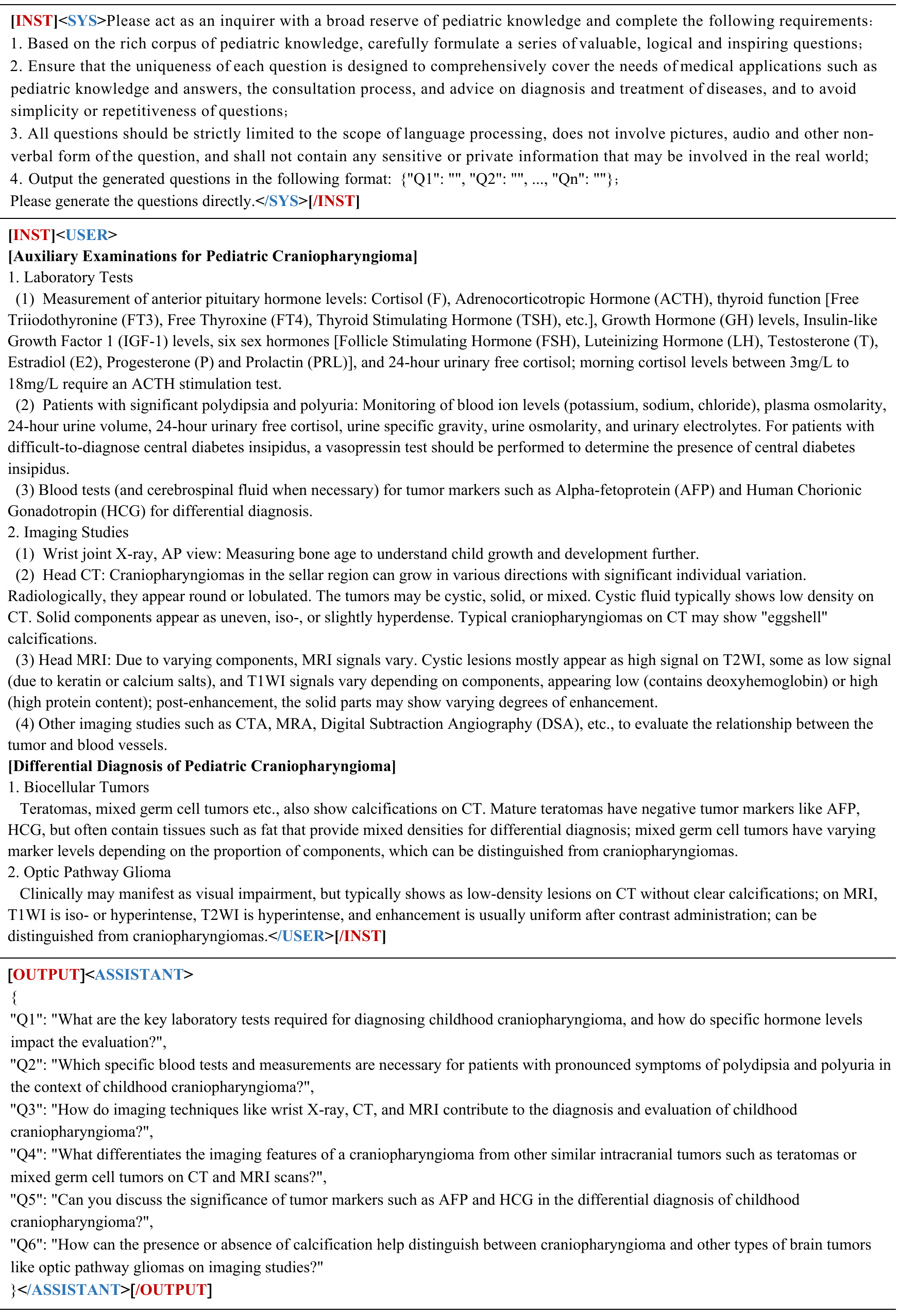}
  \caption{Illustration of the GPT-4 prompt template for the ``inquirer'' role. The original Chinese content is translated into the corresponding English version to facilitate intuitive understanding.
  }
  \label{app_sec_1_1}
\end{figure}

\begin{figure}[h]
  \centering
  \includegraphics[width=\linewidth]{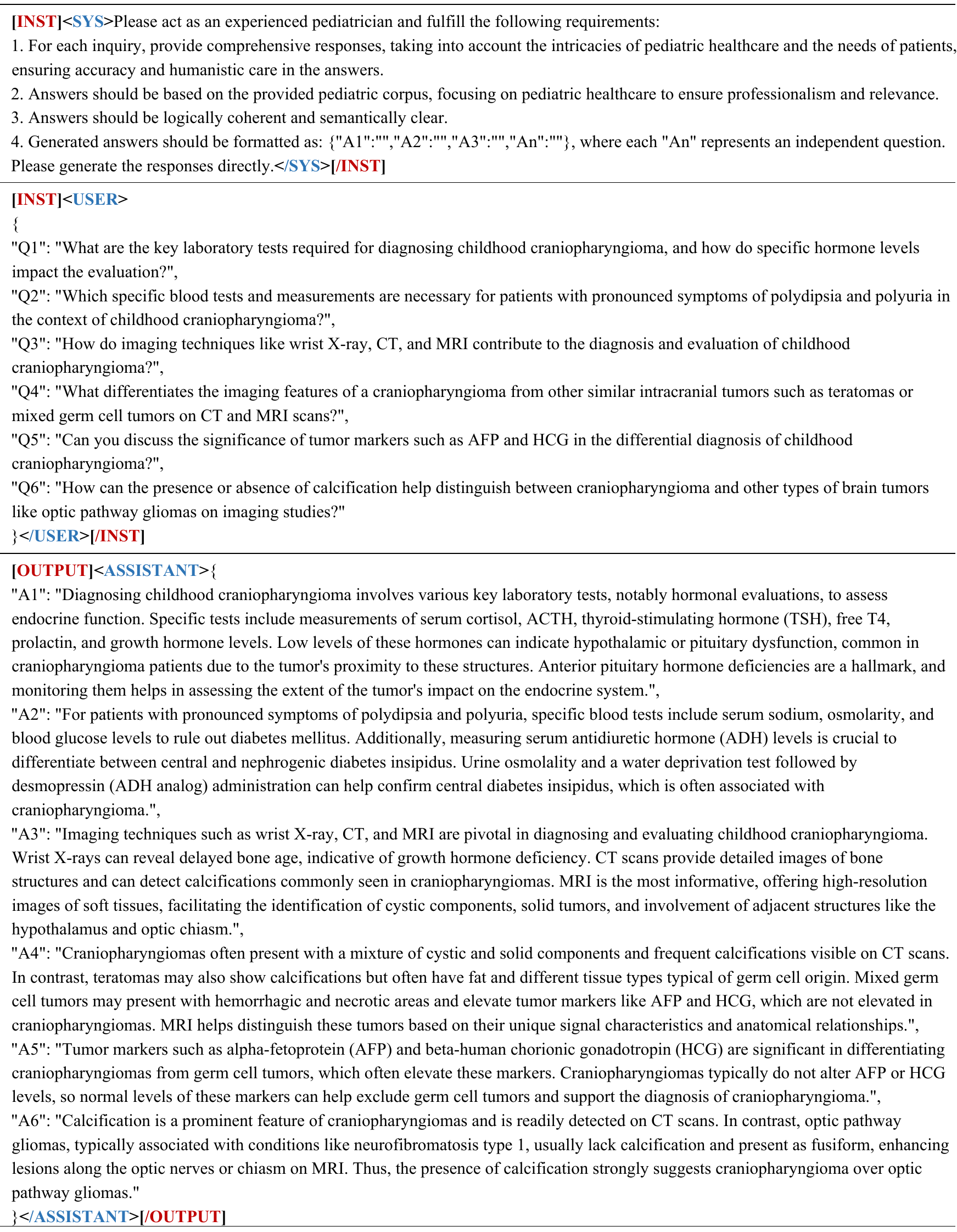}
  \caption{Illustration of the GPT-4 prompt template for the ``expert pediatrician'' role. The original Chinese content is translated into the corresponding English version to facilitate intuitive understanding.
  }
  \label{app_sec_1_2}
\end{figure}

After integrating pediatric textbooks, guidelines, and knowledge graphs into a consolidated textual database, the content is segmented according to individual diseases. Subsequently, two instances of the GPT-4 model are deployed, designated as the ``inquirer'' and the ``expert pediatrician'' respectively. Disease-specific segments are then fed into the ``inquirer'' GPT-4, tasked with formulating a series of relevant and scholarly pediatric inquiries. Following this, the original disease segments and the formulated inquiries are fed into the ``expert pediatrician'' GPT-4 to generate precise responses for each inquiry, leveraging the segmented text as the contextual reference.
We show the prompt templates for the ``inquirer'' and ``expert pediatrician'' in Figures~\ref{app_sec_1_1} and~\ref{app_sec_1_2}, respectively.

\subsection{Vanilla Doctor-patient Conversation Regularization}
\label{app_sec:1_2}

\begin{figure}[t]
  \centering
  \includegraphics[width=\linewidth]{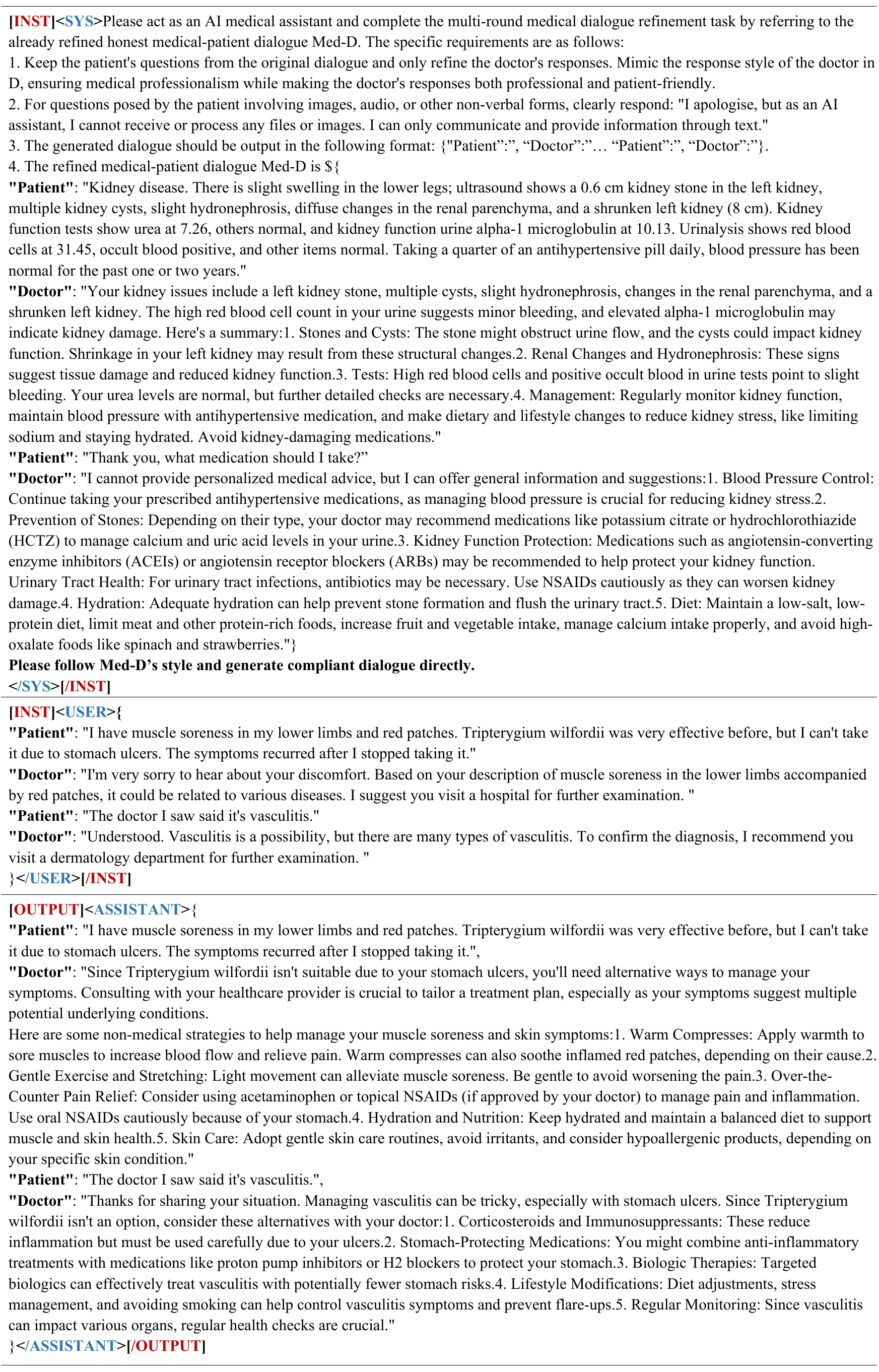}
  \caption{Illustration of the GPT-4 prompt template for vanilla doctor-patient conversation regularization. The original Chinese content is translated into the corresponding English version to facilitate intuitive understanding.
  }
  \label{app_sec_1_3}
\end{figure}

We guide GPT-4 to regularize concise and noisy doctor responses in authentic doctor-patient consultations by the context learning strategy.
Specifically, we manually craft 100 instruction examples with high-quality content to allow GPT-4 to learn doctor-like and patient-friendly behavioral styles.
In each round of regularization, we randomly sample 10 out of 100 examples to perform 10-shot context prompts. 
Immediately after that, vanilla dialogues are fed to GPT-4 as seed instructions to optimize instructions according to user requirements. 
Constrained by the space, we show the prompt case with one example in Figure~\ref{app_sec_1_3}.

\begin{figure}[t]
  \centering
  \includegraphics[width=\linewidth]{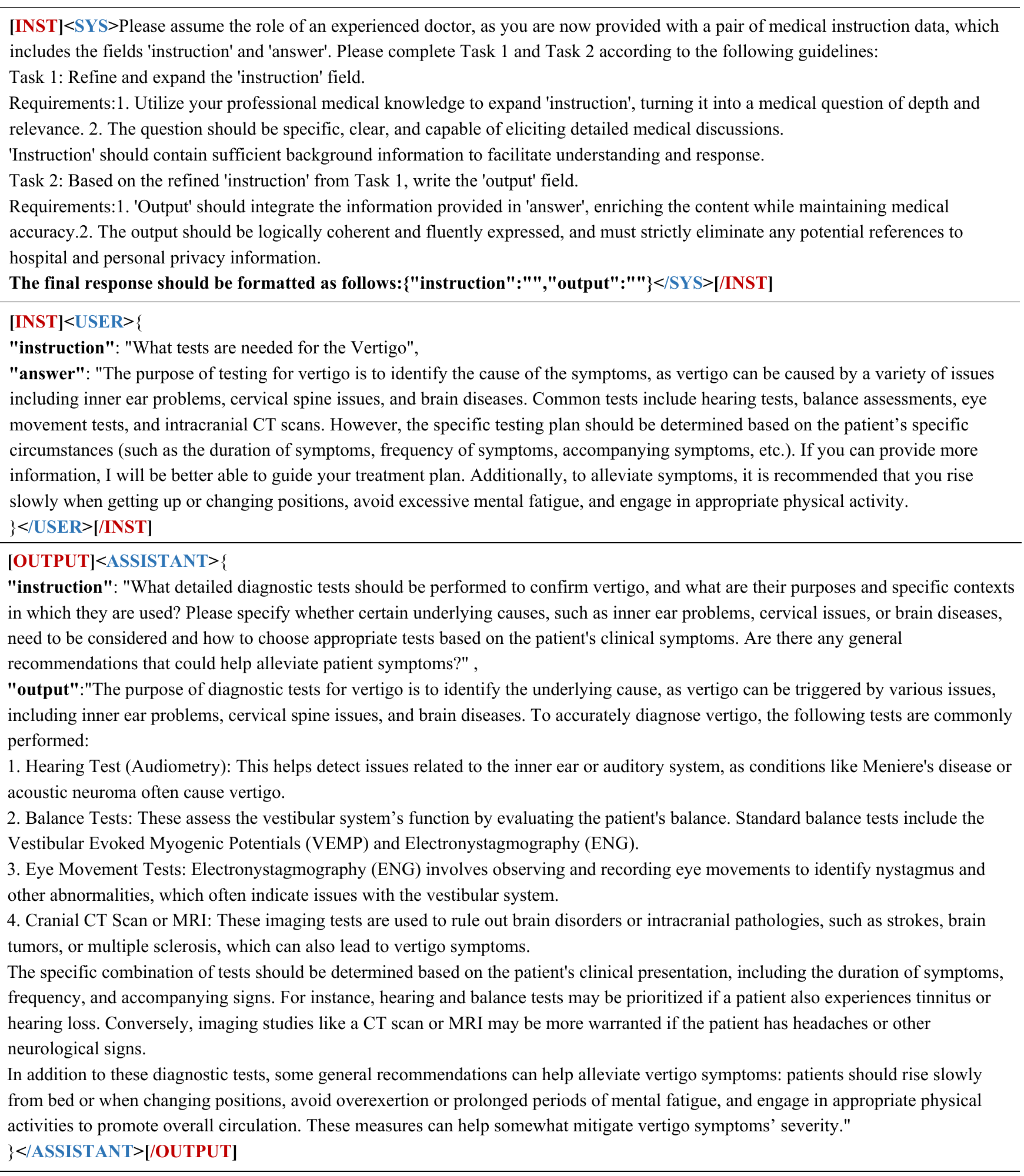}
  \caption{Illustration of the GPT-4 prompt template for progressive instruction reconstruction rule. The original Chinese content is translated into the corresponding English version to facilitate intuitive understanding.
  }
  \label{app_sec_1_4}
\end{figure}

\subsection{Progressive Instruction Reconstruction Rule}
\label{app_sec:1_3}

Medical knowledge integration from existing datasets is common but frequently imprecise, resulting in unclear or incomplete instructions and potentially inaccurate outputs. Consequently, 107,177 instructions are selected from three significant benchmarks, prioritizing quality over quantity. In this case, we design a progressive instruction reconstruction rule to refine these instructions, ensuring informative and logical model responses.

As shown in Figure~\ref{app_sec_1_4}, we first prompt GPT-4 to take the perspective of the experienced doctor to complete Tasks 1 and 2 in the given instruction and answer scenarios. Task 1 focuses on bridging the gaps in the vanilla instructions and reinforcing the completeness, professionalism, and medical relevance. Based on the refined instructions, Task 2 requires the GPT-4 to make further targeted improvements to the answer parts. In practice, this progressive reconstruction rule can activate better instruction following capabilities in advanced language models.

\section{Knowledge-enhanced Prompt}
\label{app_sec:2} 
\begin{figure}[t]
  \centering
  \includegraphics[width=\linewidth]{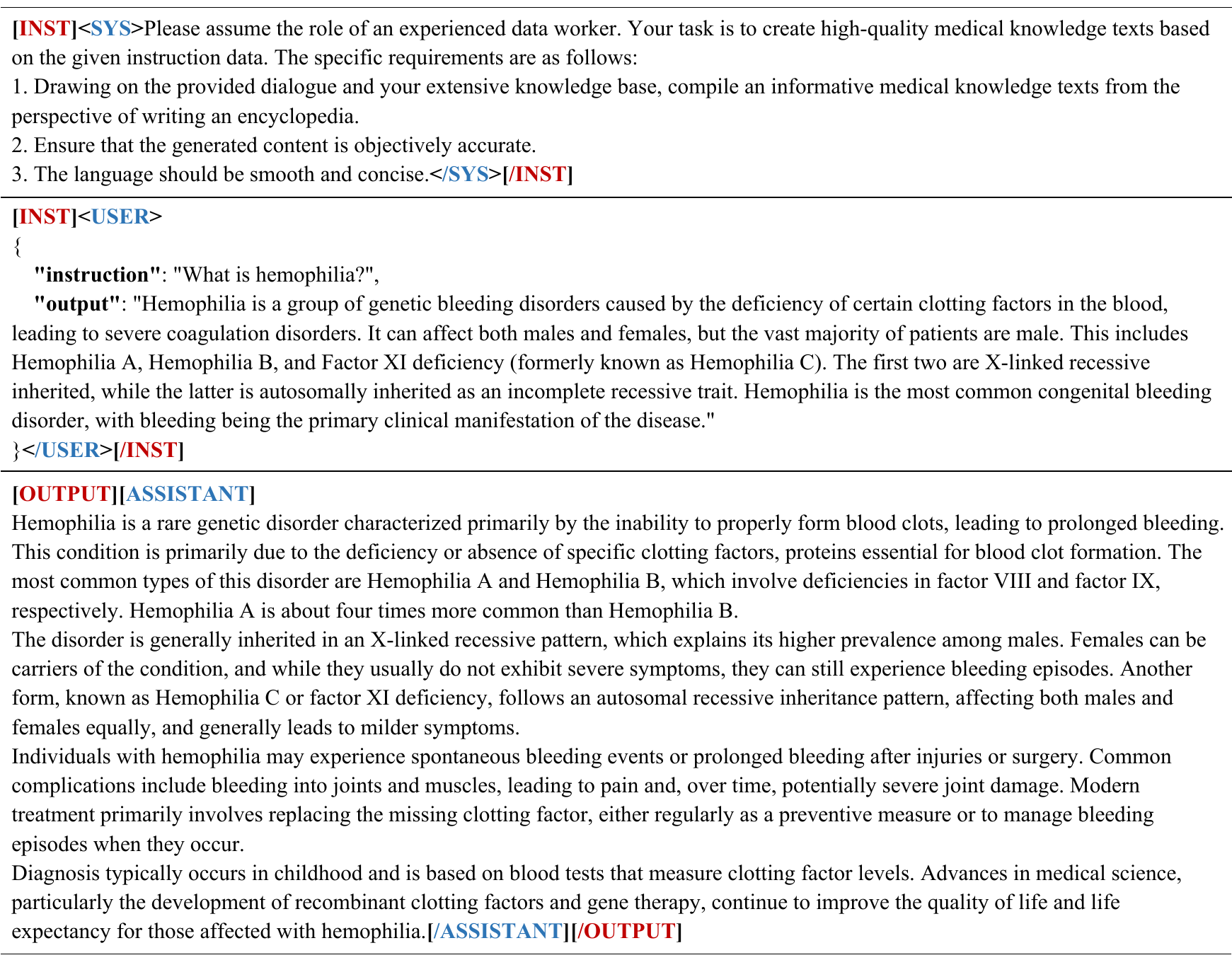}
  \caption{Illustration of the GPT-4 prompt template for improving corpus density and breadth. The original Chinese content is translated into the corresponding English version to facilitate intuitive understanding.}
  \label{app_sec_1_5}
\end{figure}

To enrich the density and breadth of the multiple-department CPT corpus, we transform the structured instruction data from the vanilla PedCorpus dataset into comprehensive medical knowledge texts using knowledge-enhanced prompts. The medical knowledge texts are integrated as complementary content to construct the PedCorpus-CPT dataset.
The prompt template is shown in Figure~\ref{app_sec_1_5}.

\section{Training Details}
\label{app_sec:3} 
In this section, we list in detail the hyper-parameter configurations for the different training phases.

\textbf{Continuous Pre-training.} During this procedure, we train each model for just a single epoch, setting the learning rate at 1e-6 and the batch size at 128. We adopt a maximum cutoff length of 4096, enabling the model to process extensive text sequences in one batch. This significantly enhances the model's contextual understanding and coherence.

\textbf{Full-parameter Supervised Fine-tuning.} In this configuration, we train all models for three epochs with a learning rate adjusted to 5e-5 and a batch size of 64, capping the maximum sequence length at 2048. We introduce a warmup\_steps setting at 200 to gradually ramp up the learning rate from an initial lower value, aiding the optimizer in adapting to gradient changes. This approach boosts stability and performance and guides the model towards a better convergence path. Also, we specify eval\_steps at 100 and save the best-performing weights on the validation set to ensure optimal results.

\textbf{Human Preference Alignment.} In this setup, we train five epochs with the learning rate set to 1e-6 and the batch size maintained at 64. To enhance the robustness and smoothness of preference learning, we adjust the control parameter $\beta$ to 0.1 and the scaling coefficient $\mu$ to 1.0. We specify eval\_steps at 100, selecting the best-performing weights on the validation set.

\textbf{LoRA-based Parameter-efficient SFT.} Here, we train three epochs with a learning rate of 1e-6 and adjust the batch size to 32. We configure the LoRA parameters by setting the rank $r$ to 8, $\alpha$ to 16, and the Dropout rate to 0.05, targeting all modules.
The default number of LoRA adapters is set to 4, including one constant universal expert and three specific experts.
Ultimately, we select the adapters that perform best on the validation set.

\section{GPT-4 Evaluation Details}
\label{app_sec:4} 
We consider four complementary dimensions in the automated evaluation to guide GPT-4 in judging the quality of model responses from a comprehensive perspective. The full definitions of these dimensions are shown as follows.

\textbf{Usefulness}: measures the extent to which the model response has pediatric expertise and relevance to the instruction intention.

\textbf{Correctness}: measures the extent to which harmful, misleading, and inaccurate information is present in the model response.

\textbf{Consistency}: measures the degree to which the model response is logically self-contradictory and the information is coherent in context.

\textbf{Smoothness}: measures whether the response content is fluent, natural, and conforms to the language expression style of human habits.

\begin{figure}[t]
  \centering
  \includegraphics[width=\linewidth]{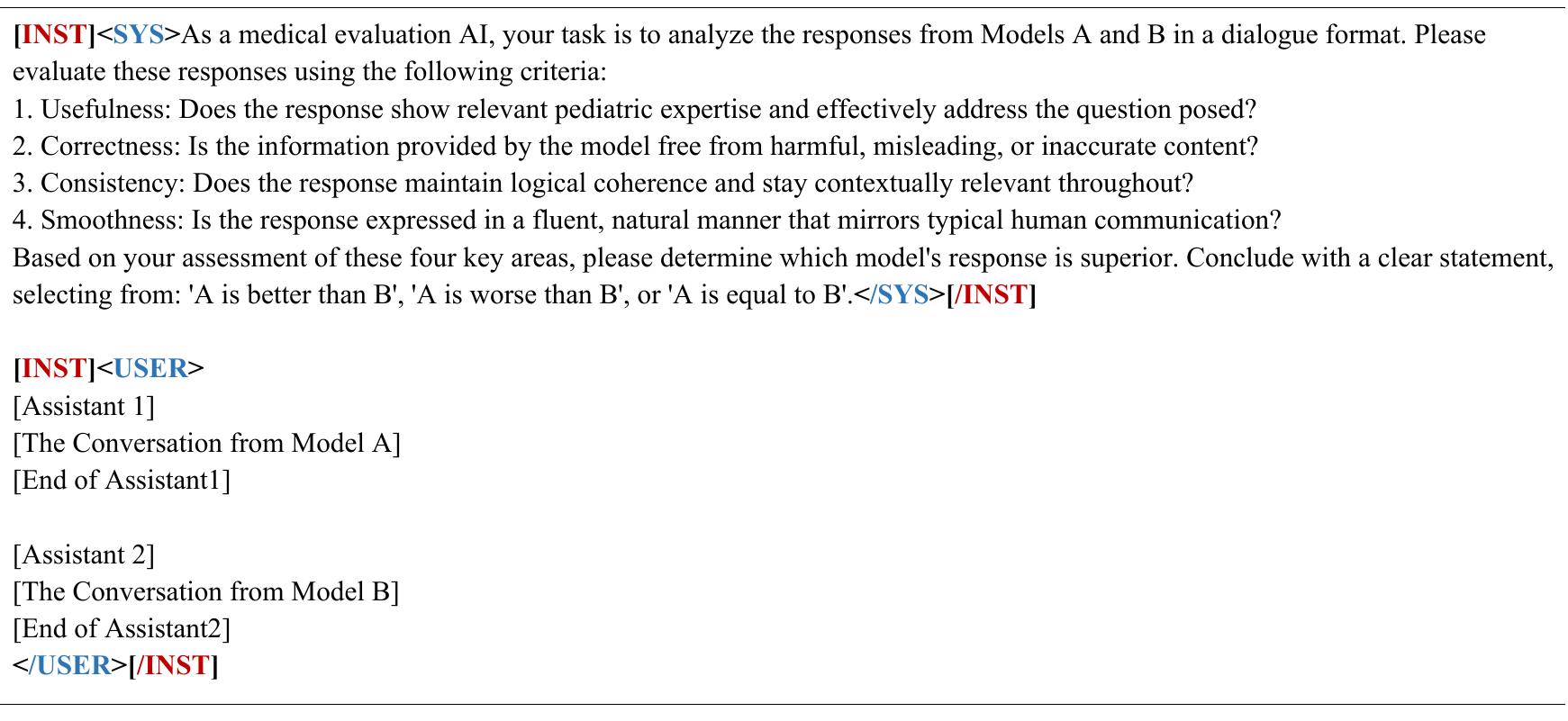}
  \caption{Illustration of the GPT-4 prompt template for evaluating paired model responses.
  }
  \label{tab_gpt_eval}
\end{figure}

In this case, we present GPT-4 with paired responses from different models, assessing various criteria such as pediatric expertise in the responses, presence of harmful, misleading, or inaccurate information, logical consistency, and the fluency and naturalness of the language, which should conform to human linguistic habits. GPT-4 assesses these responses on their merits and selects the superior one. To maintain fairness and mitigate potential position bias, the order of the responses is randomised. This methodology is supported by recent studies demonstrating GPT-4’s strong agreement with human judgment in evaluating responses.
Figure~\ref{tab_gpt_eval} demonstrates the prompt template used to evaluate the quality of paired model responses.

\vspace{-6pt}
\section{Comparison Results of Model Responses}
\label{app_sec:5} 
\begin{figure}[t]
  \centering
  \includegraphics[width=\linewidth]{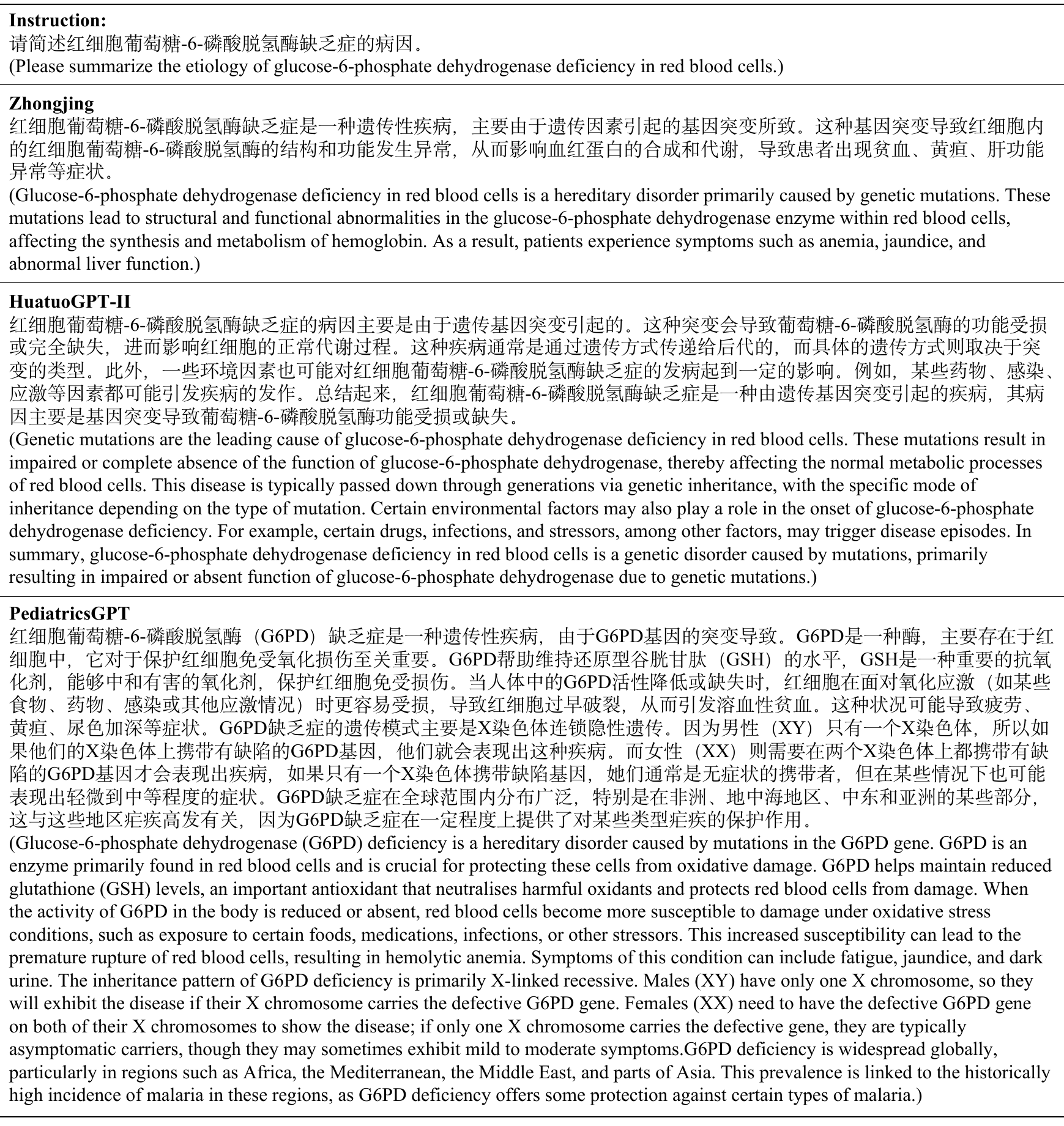}
  \caption{Comparison results of model responses on the medical knowledge question-answer task.
  }
  \label{tab_medqa}
\end{figure}

\begin{figure}[t]
  \centering
  \includegraphics[width=\linewidth]{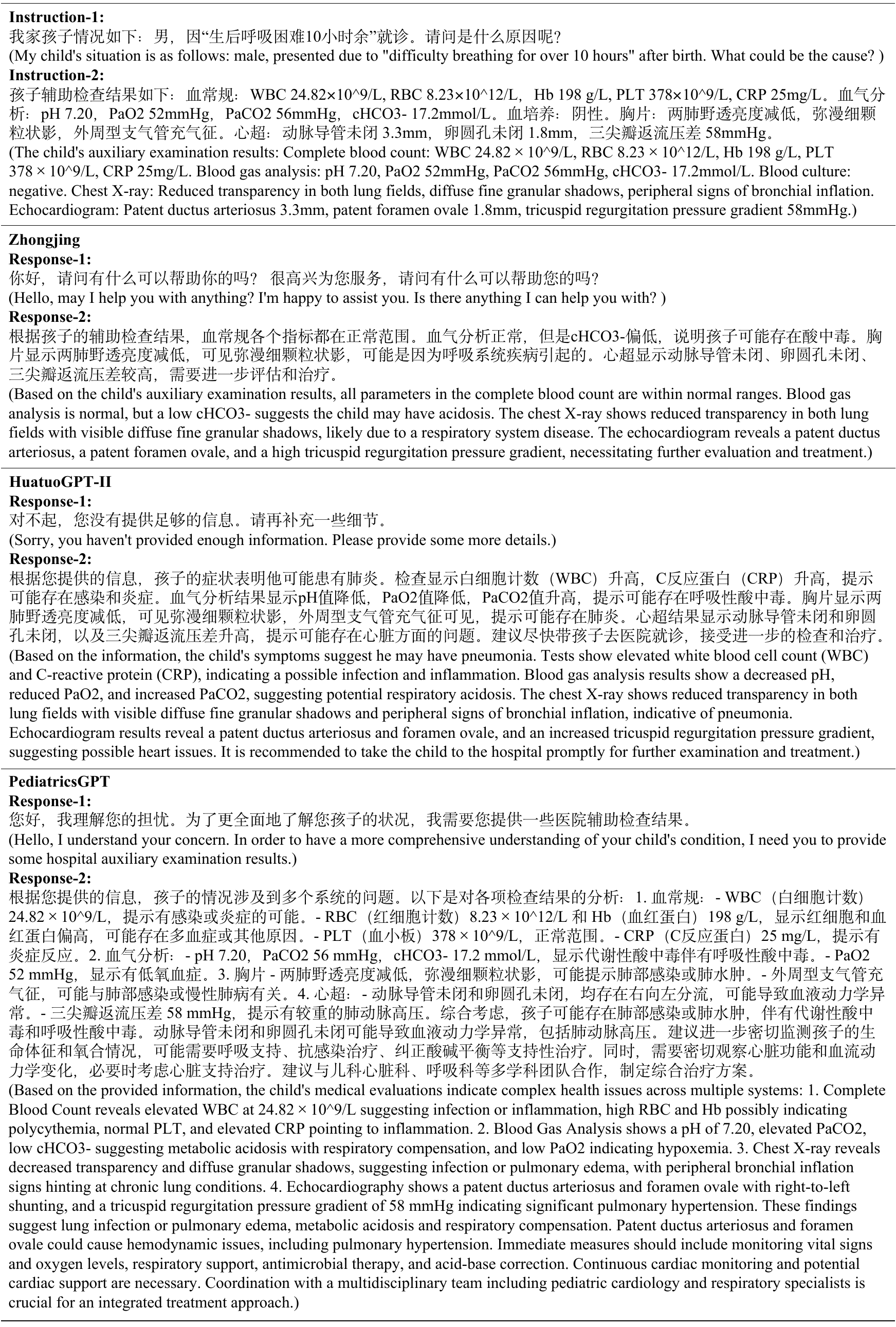}
  \caption{Comparison results of model responses on the evidence-based diagnosis task.
  }
  \label{tab_diag}
\end{figure}

\begin{figure}[t]
  \centering
  \includegraphics[width=\linewidth]{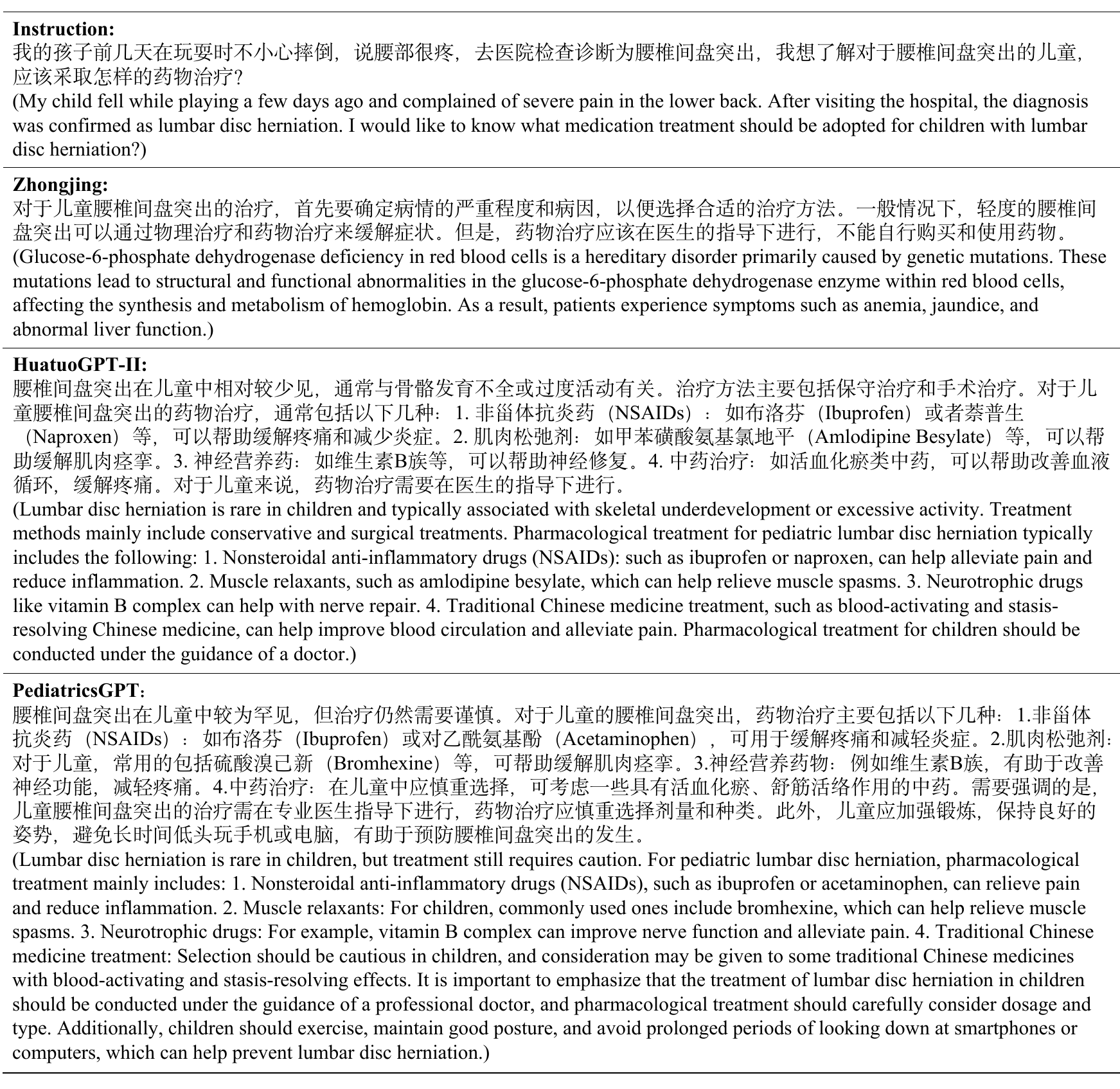}
  \caption{Comparison results of model responses on the treatment recommendation task. 
  }
  \label{tab_recom}
\end{figure}

\vspace{-6pt}
In this section, we visualize the responses of the proposed PediatricsGPT-13B and two SOTA Chinese medical LLMs across three tasks from the same medical inquiries to provide intuitive qualitative comparisons.
Specifically, Figures~\ref{tab_medqa} and \ref{tab_recom} illustrate the medical knowledge question-answer and treatment recommendation tasks, respectively, which follow a single-turn dialogue pattern. The multi-turn conversation pattern is considered in the evidence-based diagnosis task from Figure~\ref{tab_diag}.

\section{Future Work}
\label{app_sec:8} 

We list future work below to provide potential optimization directions.

\textbf{Enhancing Security Against Model Manipulation.} To mitigate the security risks associated with online deployment, our future strategy involves implementing multi-layered security measures for the proposed PediatricsGPT model. This will include advanced input validation techniques to detect and neutralize potentially malicious inputs that could manipulate model outputs. Continuous updates and patches will also be prioritized to address emerging security threats and vulnerabilities.

\textbf{Expanding Language Support.} To overcome the challenge of incomplete language coverage, we are committed to broadening the linguistic capabilities of PediatricsGPT. This expansion will involve training the model on a more diverse dataset that includes a broader range of languages and dialects, particularly those prevalent in underserved regions. By doing so, we aim to make the model more accessible and useful to a global audience, ensuring that non-Chinese speakers also benefit from reliable and localized medical information.

\clearpage

\clearpage

\clearpage
\clearpage

\newpage
\section*{NeurIPS Paper Checklist}

\begin{enumerate}

\item {\bf Claims}
    \item[] Question: Do the main claims made in the abstract and introduction accurately reflect the paper's contributions and scope?
    \item[] Answer: \answerYes{} 
    \item[] Justification: The Abstract and Section 1 show our paper's contributions and scopes.

\item {\bf Limitations}
    \item[] Question: Does the paper discuss the limitations of the work performed by the authors?
    \item[] Answer: \answerYes{} 
    \item[] Justification: We fully discuss the limitations in Section 5.

\item {\bf Theory Assumptions and Proofs}
    \item[] Question: For each theoretical result, does the paper provide the full set of assumptions and a complete (and correct) proof?
    \item[] Answer: \answerNA{} 
    \item[] Justification: This paper focuses on providing results based on metrics, GPT-4 and doctor evaluations. The related results are presented in Sections 4.3 through 4.6.

    \item {\bf Experimental Result Reproducibility}
    \item[] Question: Does the paper fully disclose all the information needed to reproduce the main experimental results of the paper to the extent that it affects the main claims and/or conclusions of the paper (regardless of whether the code and data are provided or not)?
    \item[] Answer: \answerYes{} 
    \item[] Justification: We disclose the implementation details of reproducing the results of the paper in Section 4.1 and Appendix C. 

\item {\bf Open access to data and code}
    \item[] Question: Does the paper provide open access to the data and code, with sufficient instructions to faithfully reproduce the main experimental results, as described in supplemental material?
    \item[] Answer: \answerYes{}
    \item[] Justification: We provide the link to release the relevant available resources in the Abstract.

\item {\bf Experimental Setting/Details}
    \item[] Question: Does the paper specify all the training and test details (e.g., data splits, hyperparameters, how they were chosen, type of optimizer, etc.) necessary to understand the results?
    \item[] Answer: \answerYes{}
    \item[] Justification:  We provide detailed hyper-parameter configurations in Appendix C. Also, Section 4.1 provides dataset splits and other implementation details.

\item {\bf Experiment Statistical Significance}
    \item[] Question: Does the paper report error bars suitably and correctly defined or other appropriate information about the statistical significance of the experiments?
    \item[] Answer: \answerNo{}
    \item[] Justification: Performing multiple repetitive experiments in order to compute error bars is labor-intensive and has significant overhead for the large language model development.

\item {\bf Experiments Compute Resources}
    \item[] Question: For each experiment, does the paper provide sufficient information on the computer resources (type of compute workers, memory, time of execution) needed to reproduce the experiments?
    \item[] Answer: \answerYes{} 
    \item[] Justification: We provide the computational resources needed in order to reproduce the experiments in Section 4.1.
    
\item {\bf Code Of Ethics}
    \item[] Question: Does the research conducted in the paper conform, in every respect, with the NeurIPS Code of Ethics \url{https://neurips.cc/public/EthicsGuidelines}?
    \item[] Answer: \answerYes{} 
    \item[] Justification: Our research follows the NeurIPS Code of Ethics.

\item {\bf Broader Impacts}
    \item[] Question: Does the paper discuss both potential positive societal impacts and negative societal impacts of the work performed?
    \item[] Answer: \answerYes{} 
    \item[] Justification: We fully discuss the potential social impacts in Section 5.
    
\item {\bf Safeguards}
    \item[] Question: Does the paper describe safeguards that have been put in place for responsible release of data or models that have a high risk for misuse (e.g., pretrained language models, image generators, or scraped datasets)?
    \item[] Answer: \answerYes{} 
    \item[] Justification: To ensure the safe release of data, we describe measures for dataset construction in detail in Section 3.1. To ensure that the responses of the proposed model are harmless and safe, we provide adversarial instructions to control the model behaviour in Section 3.3. Meanwhile, we perform human preference optimization for the model in Section 3.4, which further strengthens the safety and robustness of the model.

\item {\bf Licenses for existing assets}
    \item[] Question: Are the creators or original owners of assets (e.g., code, data, models), used in the paper, properly credited and are the license and terms of use explicitly mentioned and properly respected?
    \item[] Answer: \answerYes{} 
    \item[] Justification: We provide reasonable references for the datasets and models used in Sections 4.1 and 4.2, respectively.

\item {\bf New Assets}
    \item[] Question: Are new assets introduced in the paper well documented and is the documentation provided alongside the assets?
    \item[] Answer: \answerYes{} 
    \item[] Justification: We provide the relevant documentation in the Appendix.

\item {\bf Crowdsourcing and Research with Human Subjects}
    \item[] Question: For crowdsourcing experiments and research with human subjects, does the paper include the full text of instructions given to participants and screenshots, if applicable, as well as details about compensation (if any)? 
    \item[] Answer: \answerYes{}
    \item[] Justification: We offer \$300 each to participating experts. Expert doctors are asked to evaluate the quality of the models' responses. The relevant descriptions can be found in Section 4.3.

\item {\bf Institutional Review Board (IRB) Approvals or Equivalent for Research with Human Subjects}
    \item[] Question: Does the paper describe potential risks incurred by study participants, whether such risks were disclosed to the subjects, and whether Institutional Review Board (IRB) approvals (or an equivalent approval/review based on the requirements of your country or institution) were obtained?
    \item[] Answer: \answerYes{}
    \item[] Justification: The research in the paper has no risks to be faced by participants who are only used as evaluators. Moreover, we underwent an internal ethical review by the ethical review board of the partnering medical institutions with license and approval. The relevant descriptions can be found in Section 5.

\end{enumerate}

\end{document}